\documentclass[review]{elsarticle}

\usepackage{hyperref}

\usepackage{xcolor}
\usepackage{amssymb}
\usepackage{graphicx}
\usepackage{algorithmic}
\usepackage{algorithm}
\usepackage{url}
\usepackage{mathptmx} 
\usepackage{times} 
\usepackage{amsmath} 
\usepackage{amssymb}
\usepackage{hyperref}
\usepackage[flushleft]{threeparttable}
\usepackage[T1]{fontenc}
\usepackage{color}
\usepackage{enumerate}
\usepackage{wrapfig}
\usepackage{array}
\usepackage{tikz}
\usepackage{pgfplots}
\pgfplotsset{compat=1.15}
\usepackage{siunitx}

\definecolor{red}{rgb}{1.00,0.00,0.00}
\definecolor{blue}{rgb}{0.00,0.00,1.00}
\definecolor{green}{rgb}{0.1,0.50,0.1}
\definecolor{yellow}{rgb}{0.5,0.5,0.0}
\definecolor{white}{rgb}{1,1,1}

\journal{Journal of \LaTeX\ Templates}

\begin{document}

\begin{frontmatter}

\title{Robust Biped Locomotion Using Deep Reinforcement Learning on Top of an Analytical Control Approach }


\author[mymainaddress]{{Mohammadreza Kasaei}\corref{mycorrespondingauthor}}
\ead{Mohammadreza@ua.pt}
\author[mysecondaryaddress]{{Miguel Abreu}}
\author[mymainaddress]{{Nuno Lau}}
\author[mymainaddress]{{Artur Pereira}}
\author[mysecondaryaddress]{{Luis Paulo Reis}}

\address[mymainaddress]{IEETA / DETI University of Aveiro 3810-193 Aveiro, Portugal}
\address[mysecondaryaddress]{University of Porto, LIACC/FEUP, Artificial Intelligence and Computer Science Lab, Faculty of Engineering of the University of Porto, Portugal}

\begin{abstract}

This paper proposes a modular framework to generate robust biped locomotion using a tight coupling between an analytical walking approach and deep reinforcement learning. This framework is composed of six main modules which are hierarchically connected to reduce the overall complexity and increase its flexibility. The core of this framework is a specific dynamics model which abstracts a humanoid's dynamics model into two masses for modeling upper and lower body. This dynamics model is used to design an adaptive reference trajectories planner and an optimal controller which are fully parametric. Furthermore, a learning framework is developed based on Genetic Algorithm~(GA) and Proximal Policy Optimization~(PPO) to find the optimum parameters and to learn how to improve the stability of the robot by moving the arms and changing its center of mass~(COM) height. A set of simulations are performed to validate the performance of the framework using the official RoboCup 3D League simulation environment. The results validate the performance of the framework, not only in creating a fast and stable gait but also in learning to improve the upper body efficiency.   
\end{abstract}


\end{frontmatter}


\section{Introduction}
Developing a robust locomotion for bipedal robots is a challenging problem which has been investigated for decades. Although several walking approaches have been proposed and walking performance has considerably improved, it still falls short of expectations in certain domains, such as speed and stability. The question is \textit{ how is it that humans can constantly change their direction when running, while keeping their stability, but humanoids cannot?} 

To find a good answer for this question, we start by reviewing recently proposed walking frameworks, consequently identifying four points of view related with the development of a fast and stable gait. In the first point of view, the fundamental framework's core is a dynamics model of the robot, based on which the walking planner and controller are designed. In this type of framework, to reduce the complexity of developing a whole body dynamics model, some constraints are considered. Based on these constraints, an abstract model is designed instead of a real whole body dynamics model~\cite{kajita1991study,kajita2003biped,kajita2010biped,shimmyo2013biped,faraji20173lp,griffin2017walking,KASAEIIROS,kasaei2019model}. It should be mentioned that several studies exist where a whole body dynamics model is developed~\cite{yamaguchi1999development,khatib2008unified,ishihara2019full}. 
In the second point of view, the core of the framework is a set of signal generators which are coupled together to generate endogenously rhythmic signals~\cite{shan2000design,lee2013generation,liu2013central,yu2013survey}. This type of framework is called Central Pattern Generator~(CPG)-based framework and is inspired by the neurophysiological studies on invertebrate and vertebrate animals~\cite{guertin2009mammalian,zhong2012neuronal,menelaou2019hierarchical}. These studies showed that rhythmic locomotion like walking, running and swimming are generated by CPGs at the spinal cord that are connected together in a particular arrangement. In this type of framework, oscillators are assigned to each limb, typically to generate the setpoints (position, torque, etc.). Most humanoid robots have more than 20 Degrees of Freedom~(DOF), therefore, adjusting the parameters of the oscillators is not only difficult but also trial-intensive~\cite{kasaei2019fast}. Moreover, there is not a straight way to adapt sensory information to the oscillators. 
In the third point of view, walking trajectories are generated based on a heuristic algorithm such as reinforcement learning~(RL), genetic algorithm~(GA), etc~\cite{shan2000design,endo2008learning,abreu2019learning}. In this type of framework, the walking trajectories will be generated after a training period which needs many samples and takes a considerable amount of time. During training, the framework tries to learn how to generate the walking trajectories, subject to an objective function.
In the fourth point of view, the framework is designed by combining the aforementioned approaches~\cite{macalpine2012design,or2010hybrid,he2014real,kasaei2017hybrid,kasaei2019fast}. This type of framework is generally known as a hybrid walking framework. It tries to leverage the different capabilities of each approach to improve the final performance. 

After studying all types of humanoid walking frameworks, to find the answer for the question raised in the beginning of this section, let us look at how a baby starts to walk. It starts by learning to stand for a few seconds. It then improves the stability after many experiments, takes a few steps, learns how to maintain equilibrium while moving; until finally, after a long process of trial and error, a robust walking behavior emerges. This process shows how a human learns from previous experiences to improve its walking performance. Based on these explanations, we believe that the ability to learn from past experiences is the most important difference between human walking and robot walking. Particularly, a robot should be able to learn how to generate efficient locomotion according to different situations~(e.g., learning to recover its balance from postural perturbations).

In the first two types of framework, the knowledge of robots is static, generally, and does not evolve from past experiences. Therefore, they need to at least re-tune the parameters to be able to adapt to new environments. In the third type of framework, the learning process typically does not consider any dynamics model and is designed based on learning from scratch, which is trial intensive and not applicable to a real robot directly. Several research groups have been exploring how to learn from previous experiences to improve stability and robustness. We believe the fourth type is the best approach to develop a robust biped locomotion framework. 

In this paper, we propose a tight coupling between analytical control approaches and machine learning~(ML) algorithms to develop a robust walking framework. Particularly, our contribution is a biped locomotion framework composed of two major components --- an analytical planner and controller; and a fully connected neural network. The former is responsible for optimally controlling the overall state of the robot based on an abstract dynamics model. It is also responsible for generating reference trajectories using dynamic planners with genetically optimized parameters and overcome uncertainties up to a certain degree. The latter component --- a fully connected network --- is optimized with reinforcement learning to control the arms residuals and the COM height of the robot, thus improving the upper body efficiency, which impacts the overall stability and speed of the robot.

The remainder of this paper is structured as follows: Section~\ref{sec:related} provides an overview of related work. In Section~\ref{sec:Dynamics},  the concept of ZMP will be used to define a specific dynamics model which is composed of two masses. Afterwards, in Section~\ref{sec:controller}, this dynamics model will be used to design an optimal controller which is able to track the walking reference trajectories, even in the presence of uncertainties. Section~\ref{sec:walking_ref} explains how the problem of generating walking reference trajectories can be decomposed into five distinct planners. In Section~\ref{sec:learning}, we will describe our learning approach and explain its structure. The overall architecture of the proposed framework will be presented in Section~\ref{sec:overall}. In Section~\ref{sec:Simulations}, three simulations scenarios will be designed to validate the performance of the proposed framework. According to the simulation results, its discussion and comparison with related work will be provided in Section~\ref{sec:Discussion}. Finally, conclusions and future research are presented in Section~\ref{sec:Conclusion}.
	
\section{Related Work}
\label{sec:related}
Several of the proposed walking frameworks are based on learning approaches to generate a stable locomotion for biped and multi-legged robots. Using ML algorithms for biped locomotion has made remarkable progress recently. These studies showed that using these algorithms on top of analytical approaches can improve robustness and performance significantly~\cite{macalpine2012design,kasaei2019fast}. In the remainder of this section, some recent proposed walking frameworks will be categorized and reviewed, focusing on those that use ML algorithms to improve their performance.

\subsection{Combination of Model-based Walking and ML algorithms}

MacAlpine et al.~\cite{macalpine2012design} designed and implemented a learning architecture to enable a humanoid soccer agent to perform omnidirectional walk. In their architecture, the overall dynamics of a humanoid robot is abstracted by a double inverted pendulum model which is parameterized to be able to learn a set of parameters for different tasks. The performance of their framework has been validated using a set of simulations that have been designed using SimSpark~\footnote{http://simspark.sourceforge.net/}, a generic physical multiagent system simulator. The simulations results showed that their framework is able to learn multiple parameter sets according to the specified tasks.

Kasaei et al.~\cite{kasaei2019fast} proposed a closed-loop model-based walking framework. Their dynamics model is composed of two masses that takes into account the lower and upper body dynamics of the robot. Based on this dynamics model, they generate walking reference trajectories and also designed an optimal controller to track these references. They showed the performance of their framework by performing a set of simulations using a simulated NAO robot in SimSpark. Moreover, they optimized the parameters using GA and showed that the maximum forward walking speed of the simulated robot reached $80.5$ cm/s.

Carpentier et al.~\cite{carpentier2016versatile} proposed a generic and efficient walking pattern generator which is able to generate dynamically consistent motions. They argued that their approach is fast enough to generate the trajectory of COM along with the angular momentum according to the given configuration of contacts while the previous step is executing. Their method has been implemented on a real HRP-2 robot to demonstrate its interest. The experiment results showed that their method is able to generate long-step walking and climbing a staircase with handrail support.

Koryakovskiy et al.~\cite{koryakovskiy2018model} proposed two approaches for combining a Nonlinear Model Predictive Control~(NMPC) with reinforcement learning to compensate model-mismatch. The first approach deals with learning a policy to compensate control actions to minimize the same performance measure as their NMPC. The second approach was focused on learning a policy based on the difference of a transition predicted by NMPC and the actual transition. They performed a set of simulations to show the feasibility of both approaches and to compare their performances. The simulation results showed that the second approach was better than the first one. Moreover, They deployed the second approach on a real humanoid robot named Robot Leo to perform squat motion to validate the performance of their approach.

\subsection{Combination of CPG-based Walking and ML algorithms}
Song et al.~\cite{song2014cpg} designed CPG-Based Control walking framework which is able to generate stable walking, even on unknown sloped surfaces. In their framework, the walking patterns are generated based on CPG theory and a PI controller is designed according to gyroscope and accelerometer information, allowing the adjustment of the upper body's tilt angle to keep the robot's stability. They performed some experiments using a real NAO humanoid robot and the results showed that the robot is able to walk successfully on unknown slopes.

Missura et al.~\cite{missura2015gradient} proposed a walking framework which bootstraps a learning algorithm with a CPG-based walk engine. Their framework is composed of a feed-forward walking pattern generator, a state estimator and a balance controller. In their framework, while the robot is walking, the balance controller adjusts the step size based on the estimated error and also learns how to improve the walking performance by adjusting the swing leg parameters. The performance of their framework has been validated using a set of experiments on a real humanoid robot. The results showed that their framework is able to keep the robot's stability even after applying a severe push.

\subsection{Combination of CPG-ZMP based walking and ML algorithms}

Massah et al.~\cite{massah2013hybrid} developed a hybrid CPG-ZMP controller to generate stable locomotion for humanoid robots. In their approach, a set of non-linear oscillators were used to generate walking trajectories and two controllers were developed to handle small and large disturbances. They optimized the walking parameters using the differential evolution~(DE) algorithm. The performance of their approach was demonstrated in the Webots robot simulator using the NAO humanoid robot.

Liu et al.~\cite{liu2016bipedal} proposed a CPG-ZMP based walking framework which is inspired by biomechanical studies on human walking. In their framework, walking reference trajectories are generated offline according to a point mass model. They used a PD controller to modify the reference walking patterns to keep the robot's stability. Moreover, their framework takes the vertical motion of the upper body into account to generate almost stretched knees. The performance of their framework has been validated using a set of experiments on a real NAO humanoid robot. The results proved the improvement of walking stability and energy efficiency.

Kasaei et al.~\cite{kasaei2017hybrid} developed a hybrid CPG-ZMP based walk engine for biped robots. Their walk engine has a hierarchical structure and it is fully parametric. They argued that this structure allows using a policy search algorithm to find the optimum walking parameters. To show this ability, they used an optimization
technique based on Contextual Relative Entropy Policy Search with Covariance Matrix Adaptation (CREPS-CMA)~\cite{abdolmaleki2016contextual} to tune the walking parameters. The performance of their walk engine engine has been validated by showing a fast and stable omnidirectional walk using a simulated Nao robot in Simspark~($59cm/s$).

\subsection{Learning to walk from scratch}

Abreu et al.~\cite{abreu2019learning} applied a reinforcement learning algorithm to develop a fast and stable running behavior from scratch. In their approach, the environment has been represented by 80 states and the action space is composed of 20 actions which were all the joints of a simulated humanoid robot. They used the Proximal Policy Optimization~(PPO) based on the implementation provided by OpenAI~\cite{baselines}. The performance of their approach was shown by learning sprinting and stopping behaviors. The results demonstrated that both behaviors are stable and the sprinting speed stabilizes around $2.5 m/s$ which was a considerable improvement.

\begin{figure}[!t]
	\begin{centering}
		\begin{tabular}	{c}	
			\includegraphics[width=0.95\linewidth,trim= 0cm 0cm 0cm 0cm,clip] {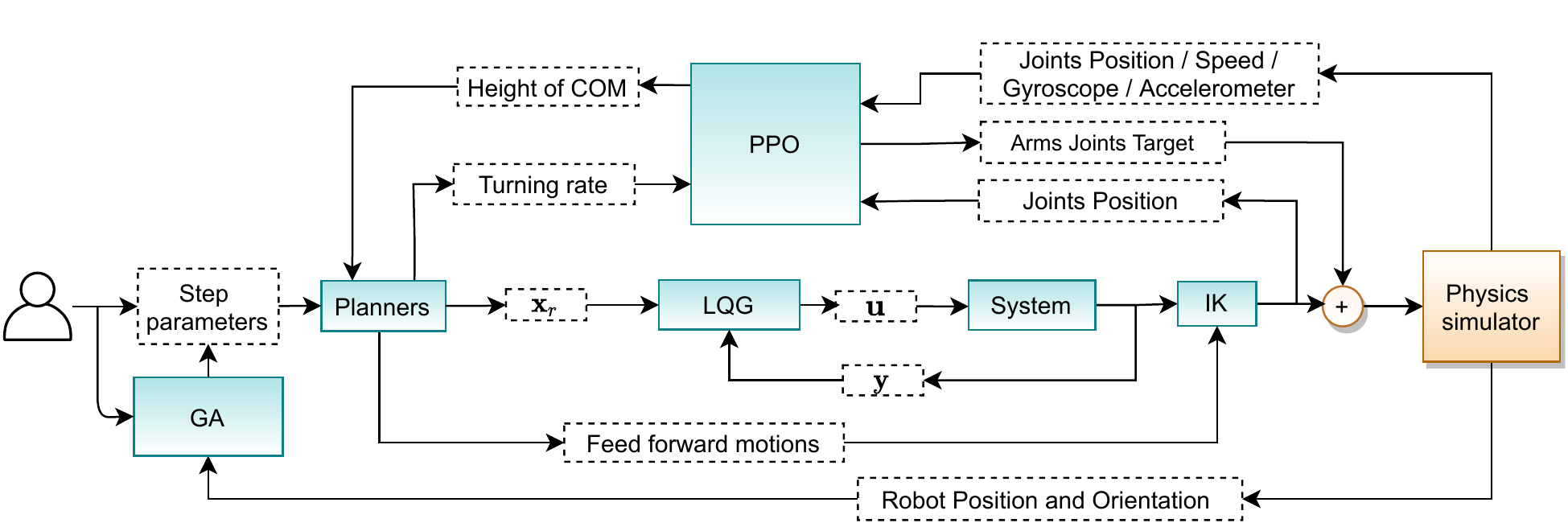}		
		\end{tabular}	
	\end{centering}			
	\caption{An abstract overview of the proposed framework. The highlighted boxes represent functional modules and the white boxes correspond to exchange data among them.  }
	\label{fig:abs_architecture}
\end{figure}

Most of the aforementioned works combine a simplified model-based or a model-free approach with ML approaches to improve the performance of their walking. In the rest of this paper, we develop an optimal closed-loop walking pattern generator based on a more complex dynamics model which takes into account the vertical motion of the COM and the torso's dynamics. Besides, we use the PPO algorithm which is one of the most successful deep reinforcement learning methods, on top of our walking pattern generator to improve its robustness and efficiency and also to provide more human-like walking. An abstract overview of the proposed framework is depicted in Figure~\ref{fig:abs_architecture}.

\section{Dynamics Model and Stability Criteria}
\label{sec:Dynamics}
When designing a model-based walking, two general perspectives exist: (i)~considering an abstract dynamics model which takes into account a trade-off between accuracy and simplicity; (ii)~considering a whole-body dynamics model which is more accurate but, not only is it platform dependent but also resource-intensive due to its non-linear nature. In the rest of this section, the concept of Zero Momentum Point~(ZMP) will be reviewed and then used to define an abstract dynamics model of a humanoid robot.
\subsection{Zero Momentum Point}
ZMP has been proposed in~\cite{vukobratovic1970stability} and is currently one of the most successful metrics in the walking literature. Particularly, it is a point on the ground where the ground reaction force~(GRF) acts to cancel the gravity and the inertia. Normal human walking is a periodic motion which can be decomposed into two main phases: (i)~Single Support~(SS) and (ii)~Double Support~(DS)~\cite{winter1990control}.
During SS phase, only one foot is in contact with the ground and the other foot swings toward the next planned foot position. In this paper, we used the ZMP as our main criterion for analysing the stability of the robot while performing walking and it can be defined using the following equation:
\begin{equation}
p_x = \frac{\sum_{k=1}^{n} m_k x_k (\ddot{z}_k + g) - \sum_{k=1}^{n} m_k z_k \ddot{x}_k  }{\sum_{k=1}^{n} m_k (\ddot{z}_k + g)} \quad , 
\label{eq:zmp}
\end{equation}
\noindent
where $n$ represents the number of parts that are considered in the dynamics model, $m_k$ is the mass of each part, $(x_k,\dot{x}_k)$, are the horizontal position and acceleration, and $(z_k,\ddot{z}_k)$ are the vertical position and acceleration of each mass, respectively. 

\subsection{Dynamics Model}
\begin{figure}[!t]
	\centering
	\begin{tabular}	{c c c}			
		\includegraphics[width=0.28\textwidth, trim= 10.5cm 2.5cm 10cm 2cm,clip] {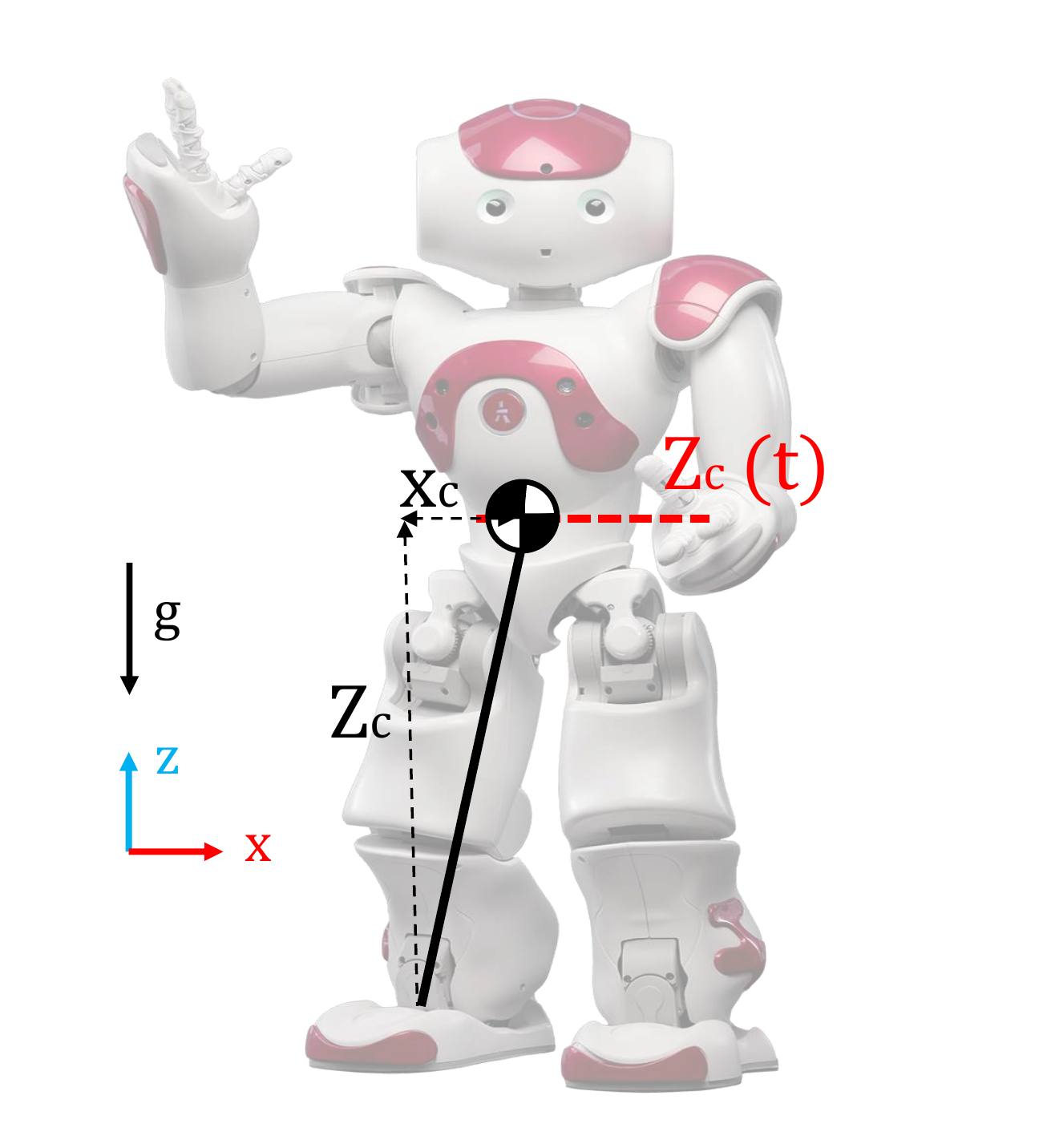} &
		\includegraphics[ width=0.28\textwidth,trim= 10.5cm 2.5cm 10cm 2cm,clip] {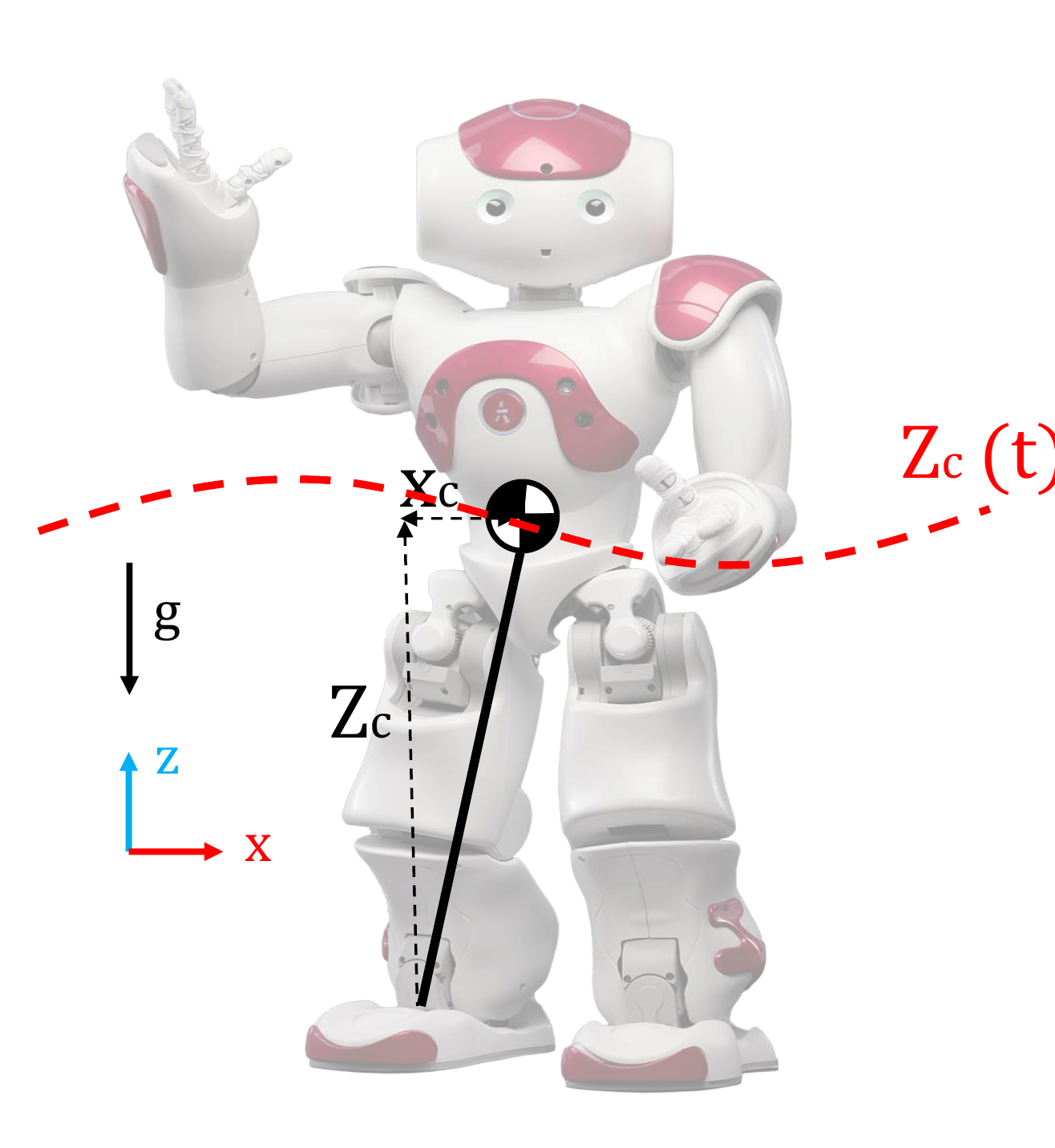}&
		\includegraphics[width=0.28\textwidth,trim= 10.5cm 2.5cm 10cm 1.5cm,clip] {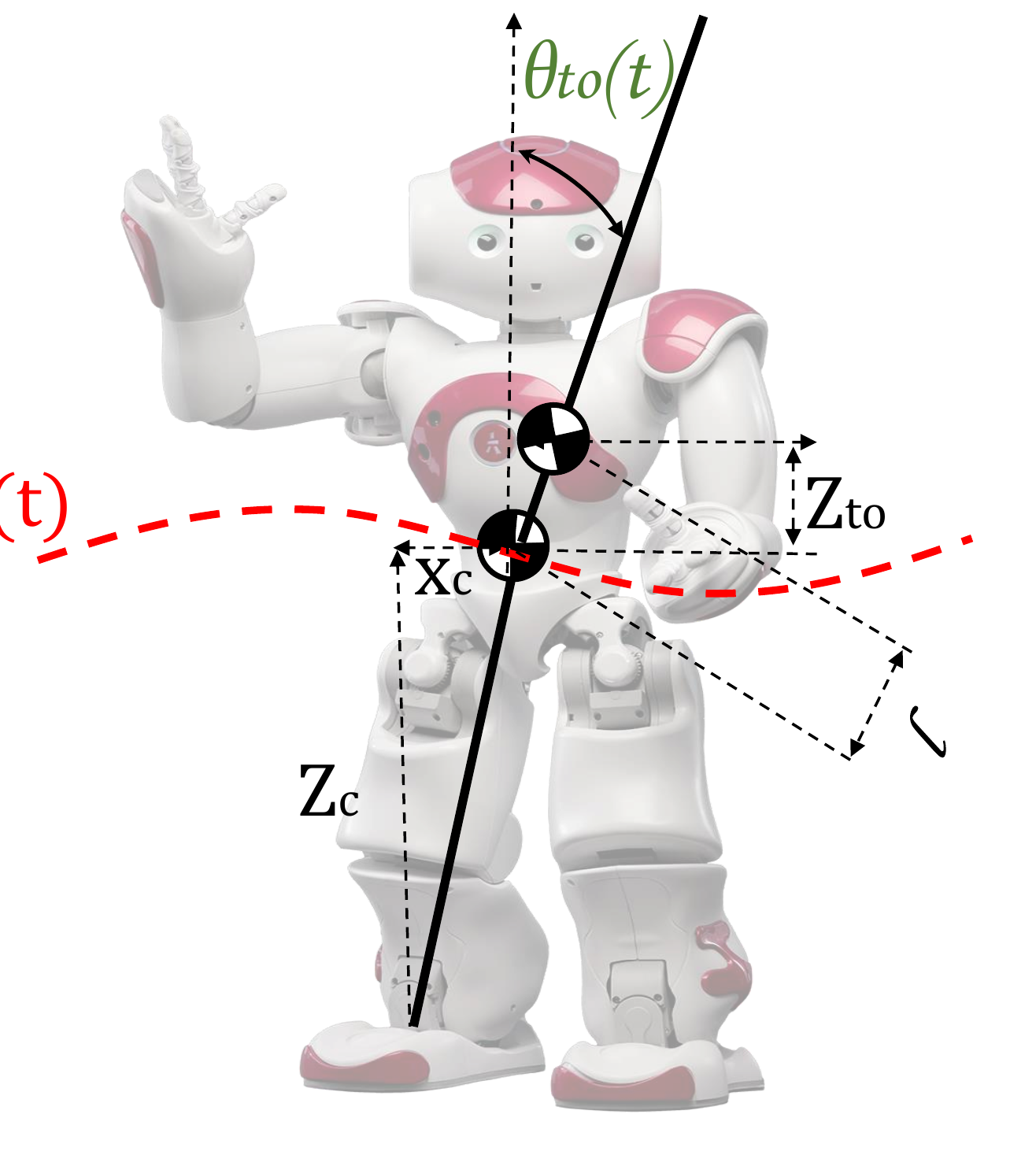}\\
		(a) & (b) & (c)
	\end{tabular}
	\vspace{-0mm}
	\caption{ Schematics of the dynamics models: (a)~LIPM; (b)~LIPM with vertical motion of COM; (c)~Proposed model.}
	\vspace{-0mm}
	\label{fig:DynamicsModel}
\end{figure}

Although considering a full body dynamics model is not impossible, it generally needs powerful computational resources. Therefore, it is not affordable for real-time implementation. To reduce the complexity of the model and its computation cost, the overall dynamics is approximated by an abstract model. Kajita and Tani~\cite{kajita1991study} proposed an abstract model named Linear Inverted Pendulum Model~(LIPM) which is a well-known abstract model in the community. LIPM is popular because it provides a simple, fast and efficient solution for walking dynamics that is suitable for real-time implementation. In this model, the overall dynamics of the robot is abstracted to a single mass that is connected to ground via a massless rod. Additionally, this model assumes that the vertical motion of the mass is restricted by a horizontally defined plane. According to these assumptions and using a set of predefined footsteps, the trajectory of Center of Mass~(COM) can be obtained from a straightforward analytical solution which guarantees long-term stability. It should be mentioned that based on these assumptions, the equations in sagittal and frontal planes are equivalent and independent, therefore we just derive the equation in the sagittal plane. The schematic of this model is depicted in Figure~\ref{fig:DynamicsModel}(a). Using~(\ref{eq:zmp}) and considering the LIPM's assumptions, the COM's motion equation can be obtained as follows:
\begin{equation}
\ddot{x}_c = \omega^2 ( x_c - p_x) \quad ,
\label{eq:lipm}
\end{equation}
\noindent
where  $\omega = \sqrt{\frac{g}{z}}$ is the natural frequency of the pendulum, $p_x$ and $x_c$ represent the positions of ZMP and COM, respectively.

As aforementioned, LIPM tries to keep the COM's vertical position at a predefined position which causes the knee joints to be always bent. Indeed, walking with bent knees consumes more energy and does not resemble human walking~\cite{kajita2019position}. To release this constraint and generate more energy efficient and human-like walking, a sinusoidal motion is assigned to the vertical motion of the COM:  
\begin{equation}
z_c = z_0 +  A_z\cos(\frac{2\pi}{StepTime} t + \phi)\quad ,
\label{eq:verticalCOM}
\end{equation}
\noindent
where $z_0$ denotes the COM's initial height, $A_z$ is the amplitude, and $\phi$ represents the phase shift of the COM's vertical sinusoidal motion. The initial value of these parameters are determined by an expert. Additionally, a controller can be designed to adjust these parameters based on sensory feedback. Although the current version of the dynamics model is able to provide fast and stable walking, it is not good enough to generate a very fast walking. In fact, in some situations like when a push is applied, the COM accelerates forward, and, as a consequence, the ZMP goes behind the Center of Gravity~(COG). In this situation, the robot tries to decelerate the COM by applying a compensating torque at its ankles, keeping the ZMP inside the support polygon. The compensating torque will be saturated once the ZMP is at the support polygon's boundary and, consequently, the robot is going to be unstable. In such situations, a human moves its torso to keep the ZMP inside the support polygon and prevent falling.

To consider the effect of the torso's motion in the dynamics model, another mass should be added to the dynamic model. This modification changes the dynamics model to be non-linear. Therefore, it does not have an analytical solution and it should be solved numerically. Biomechanical analysis of human walking showed that the torso motion can be represented by a sinusoidal function whose motion parameters are dependent on the current robot state, and terrain conditions. The interesting point is that if the torso is considered as a mass with a small sinusoidal movement relative to the hip ($\sin(\theta_{to})= \theta_{to}$), the dynamics model can keep its linearity. The schematic of this model is depicted in Figure~\ref{fig:DynamicsModel}(c) and it can be represented by the following equation: 
\begin{equation}
\begin{aligned}
\ddot{x}_c =  \mu ( x_c+\frac{\alpha l}{1+\alpha}&\theta_{to} - p_x)- \frac{\alpha\beta l}{1+\alpha\beta}\ddot{\theta}_{to} \\
\alpha = \frac{m_{to}}{m_c}, \quad \beta = \frac{z_{to}}{z_{c}}, \quad \mu =& \frac{1+\alpha}{1+\alpha\beta}\omega^2, \quad x_{to} = x_c+l\theta_{to}
\end{aligned}
\label{eq:lipm_new}
\end{equation}
\noindent
where $x_{to}$ denotes the position of torso, $\theta_{to}$, $l$ are the angle of torso and length of torso,
$m_{c}$,$m_{to}$ represent the masses of lower body and torso, $z_{to}$, $z_{c}$ are the torso and COM height, respectively.

\section{Controller}
\label{sec:controller}
In this section, an optimal controller will be designed for tracking the reference trajectories to minimize the tracking error. To do that, Equation~(\ref{eq:lipm_new}) will be represented as a linear state space system. Then, this system will be discretized to be used in a discrete-time implementation. Afterwards, we will explain how this system can be used to design an optimal controller.
\begin{figure}[!t]
	\centering
		\begin{tabular}	{c}	
			\includegraphics[width=0.8\linewidth,trim= 2.5cm 6.8cm 4.5cm 1cm,clip] {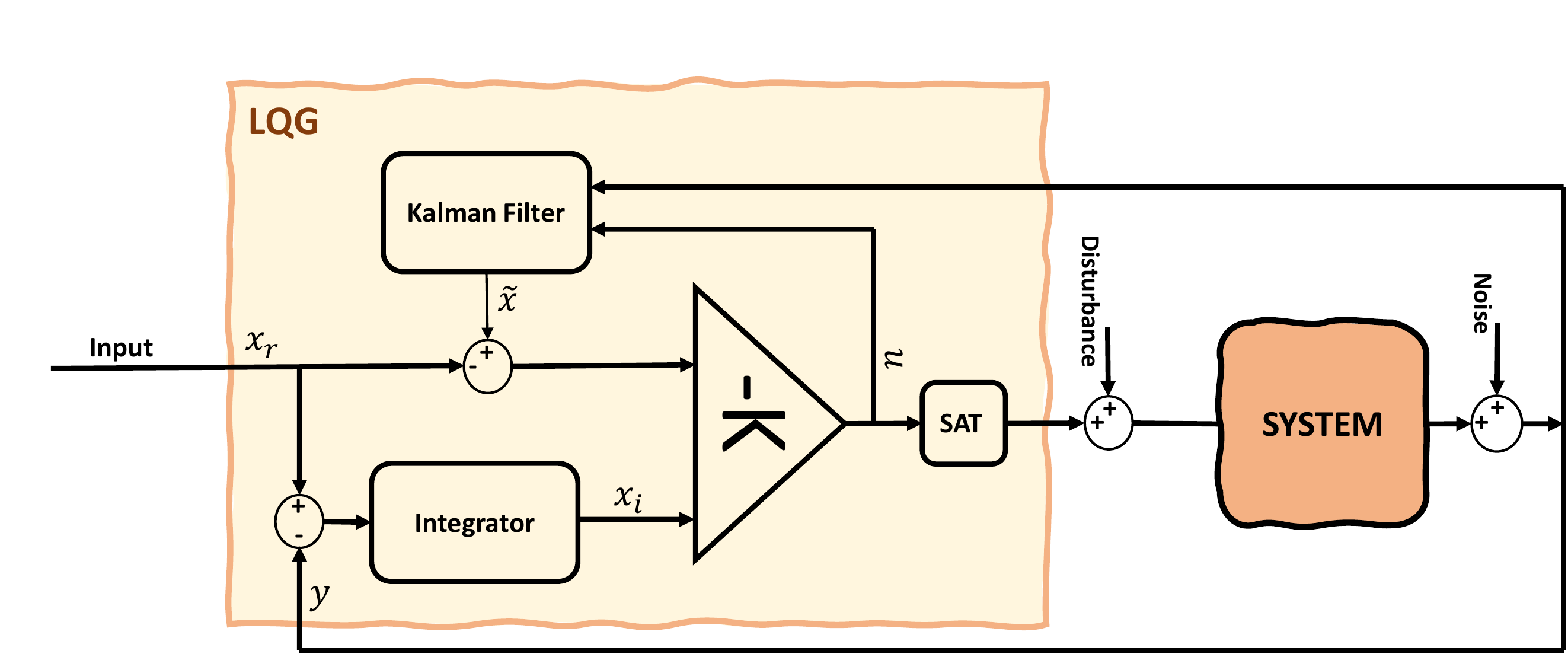}		
		\end{tabular}			
	\caption{Overall architecture of the proposed controller.}
	\label{fig:ControllerArchitecture}
\end{figure}
The process of designing the controller starts by defining a linear state space system based on Equation~\ref{eq:lipm_new}:
\begin{equation}
\label{eq:statespace_alpha}
\begin{aligned}
\frac{d}{dt} \underbrace{\begin{bmatrix} x_c\\ \dot{x}_c \\ \theta_{to} \\ \dot{\theta}_{to}\end{bmatrix}}_X
=
\underbrace{\begin{bmatrix} 
0 & 1 & 0 & 0  \\ 
\mu & 0 & \frac{\mu\alpha l}{1+\alpha} & 0  \\
0 & 0 & 0 & 1\\
0 & 0 & 0 & 0				  
\end{bmatrix}}_A
&\underbrace{\begin{bmatrix} x_c\\ \dot{x}_c \\ \theta_{to} \\ \dot{\theta}_{to}\end{bmatrix}}_X
+
\underbrace{\begin{bmatrix} 
0 & 0  \\
-\mu &\frac{-\alpha\beta l}{1+\alpha\beta}\\
0 &0\\
0 &1
\end{bmatrix}}_B
\underbrace{\begin{bmatrix} 
p_x \\ \ddot{\theta}_{to}  
\end{bmatrix}}_u\\
y =
\underbrace{\begin{bmatrix} 
	1 & 0 & 0 & 0  \\ 
	0 & 1 & 0 & 0\\
	0 & 0 & 1 & 0\\
	0 & 0 & 0 & 1				  
	\end{bmatrix}}_C
&\underbrace{\begin{bmatrix} x_c\\ \dot{x}_c \\ \theta_{to} \\ \dot{\theta}_{to}\end{bmatrix}}_X.
\end{aligned}
\end{equation}
The presented system is a continuous system and should be discretized for implementation in discrete time. To discretize this system, we assume that $\dot{x}_c, \dot{\theta}_{to}$ are linear. Therefore $p_x, \ddot{\theta}_{to}$ are constant within a control cycle. Thus, the discretized system can be represented as follows:
\begin{equation}
\begin{gathered} 
X(k+1)=A_d X(k)+B_d u(k)\quad \\
y(k) = C_d X(k)
\end{gathered} 
\label{eq:statespace_lipm_disc} 
\end{equation}
\noindent
where $k$ represents the current sampling instance, $A_d, B_d, C_d$ are the discretized versions of the $A, B, C$ matrices in~(\ref{eq:statespace_alpha}), respectively.

According to this discretized dynamics model, an optimal closed-loop controller can be designed to track the reference trajectories. This controller is a Linear-Quadratic-Gaussian~(LQG) which is composed of two main modules: a state estimator and an optimal controller gain. The overall architecture of this controller is depicted in Figure~\ref{fig:ControllerArchitecture}. In the remaining of this section, each module will be explained and the overall performance of the controller will be validated. 

\subsection{State Estimator}
An LQG controller is able to track the reference trajectories even in the presence of measurement noise. This controller uses a state estimator to cancel the effect of uncertainties which can be raised because of many aspects like errors in modeling the system, sensor noise, backlash of the gears, etc. In particular, this controller uses a state estimator to estimate the current state of the system according to the control inputs and the observations.
\begin{figure}[t]
	\center
	\begin{tabular}	{cc}		
		\includegraphics[width=0.45\linewidth]{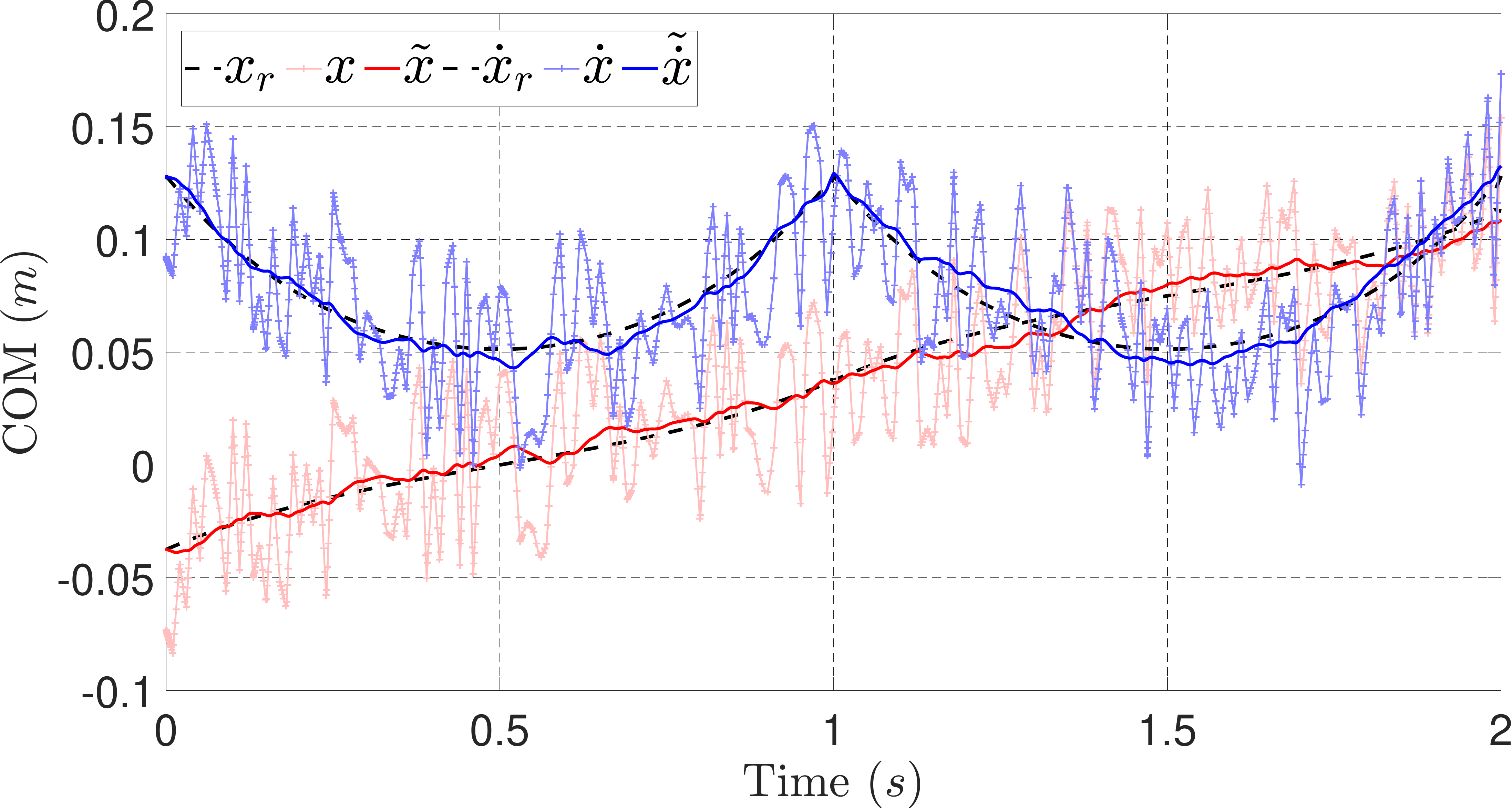}&
		\includegraphics[width=0.45\linewidth]{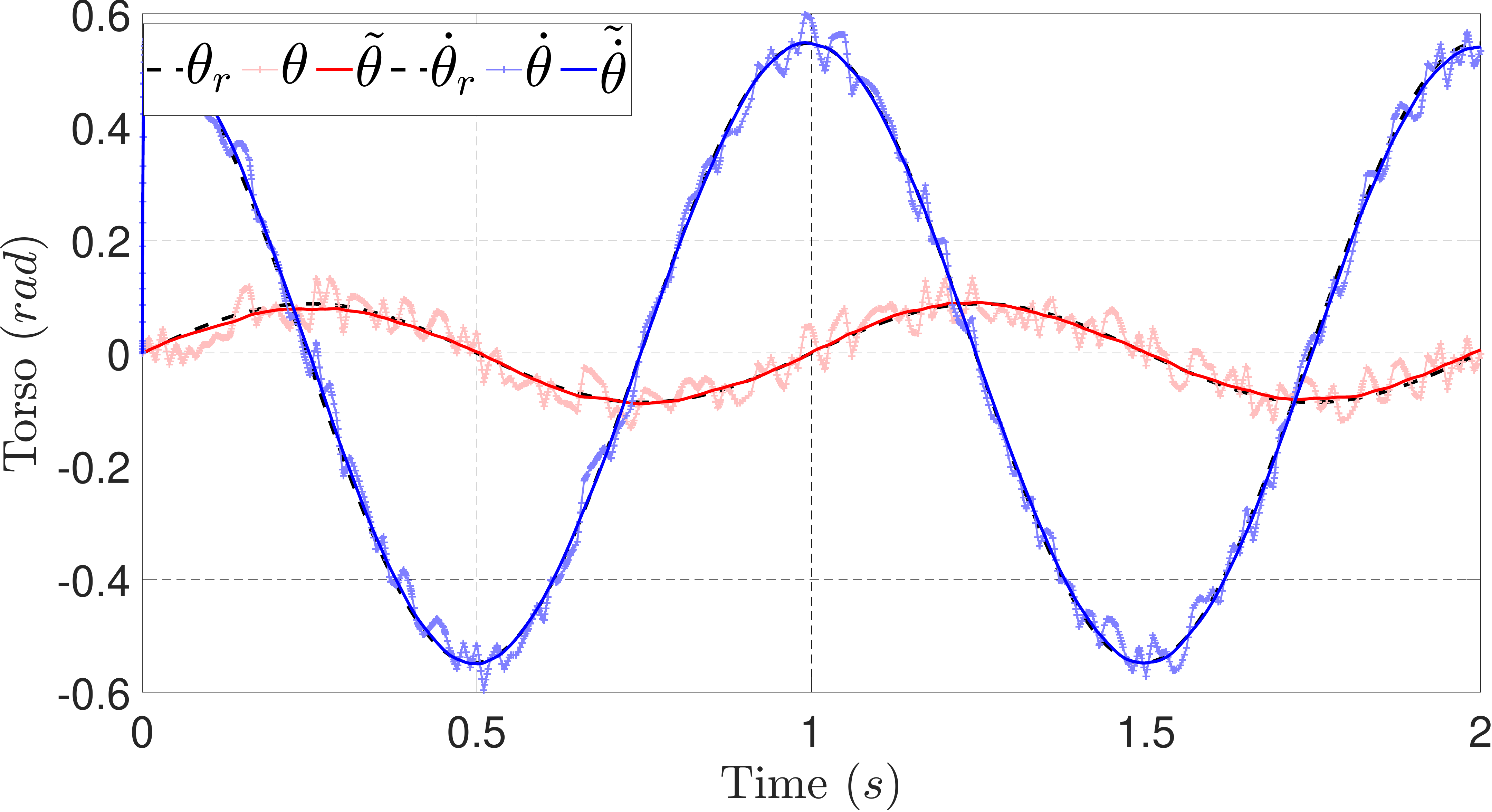}	
	\end{tabular}
    \vspace{-2mm}
	\caption{Simulation results of examining the state estimator performance. In this simulation, the measurements are affected by a Gaussian noise $\mathcal{N}(\text{0, 6.25e-4})$ to simulate uncertainties. In these plots, light-blue and light-red lines represent the measurements, solid-blue and solid-red lines are the estimated values, dashed black lines represent the references.}
	\label{fig:Kalman}       
	\vspace{-0mm}
\end{figure}

In our target framework, we considered that the position of the joints is available through measurements and the torso orientation can be obtained based on an Inertial Measurement Unit~(IMU) information which is mounted on the torso. Based on the joint information and using a Forward Kinematic~(FK) model of the robot, the current configuration of the robot can be estimated. In this estimation, the support foot is considered to be in flat contact with the ground, which is not always true. Therefore, the whole body orientation with respect to the ground should be added to this estimation. To do so, the IMU information is used to rotate the current configuration. Based on this configuration, the COM position can be estimated at each control cycle and its velocity can be obtained from its position's derivative, followed by a first-order lag filter.

To validate the performance of this module, a simulation has been designed. In this simulation, the observations are modeled as a stochastic process by applying additive Gaussian noise to the measured states. The simulation results are shown in Figure~\ref{fig:Kalman}. According to the simulation results, the state estimator is able to estimate the states perfectly.

\subsection{Optimal Gain}
As shown in Figure~\ref{fig:ControllerArchitecture}, the optimal control law is obtained using the following formulation:
\begin{equation}
u = -K
\begin{bmatrix}
\tilde{x} - x_{r}\\
x_i
\end{bmatrix} \quad ,
\end{equation}
\noindent
where $\tilde{x}$, $x_{r}$ denote the estimated states and the reference states, respectively. $x_i$ is the integration of error which is used to eliminate the steady-state error, $K$ represents the optimal gain of the controller that should be designed to minimize the following cost function:
\begin{equation}
J(u) = \int_{0}^{\infty} \{ z^\intercal Q z + u^\intercal R u \} dt \quad ,
\end{equation}
\noindent
where $z = [\tilde{x} \quad x_i]^\intercal $, $R$ and $Q$ are positive-semidefinite and positive-definite matrices which are determined by an expert. In fact, these matrices determine a trade-off between cost of control effort and tracking performance. Therefore, the performance of the controller is sensitive to these matrices. It should be noted that there is a straightforward solution to determine $K$ based on solving a differential equation named Riccati Differential Equation~(RDE).
\begin{figure}[!t]
	\centering
	\begin{tabular}	{c  c}			
		\includegraphics[width=0.47\textwidth ] {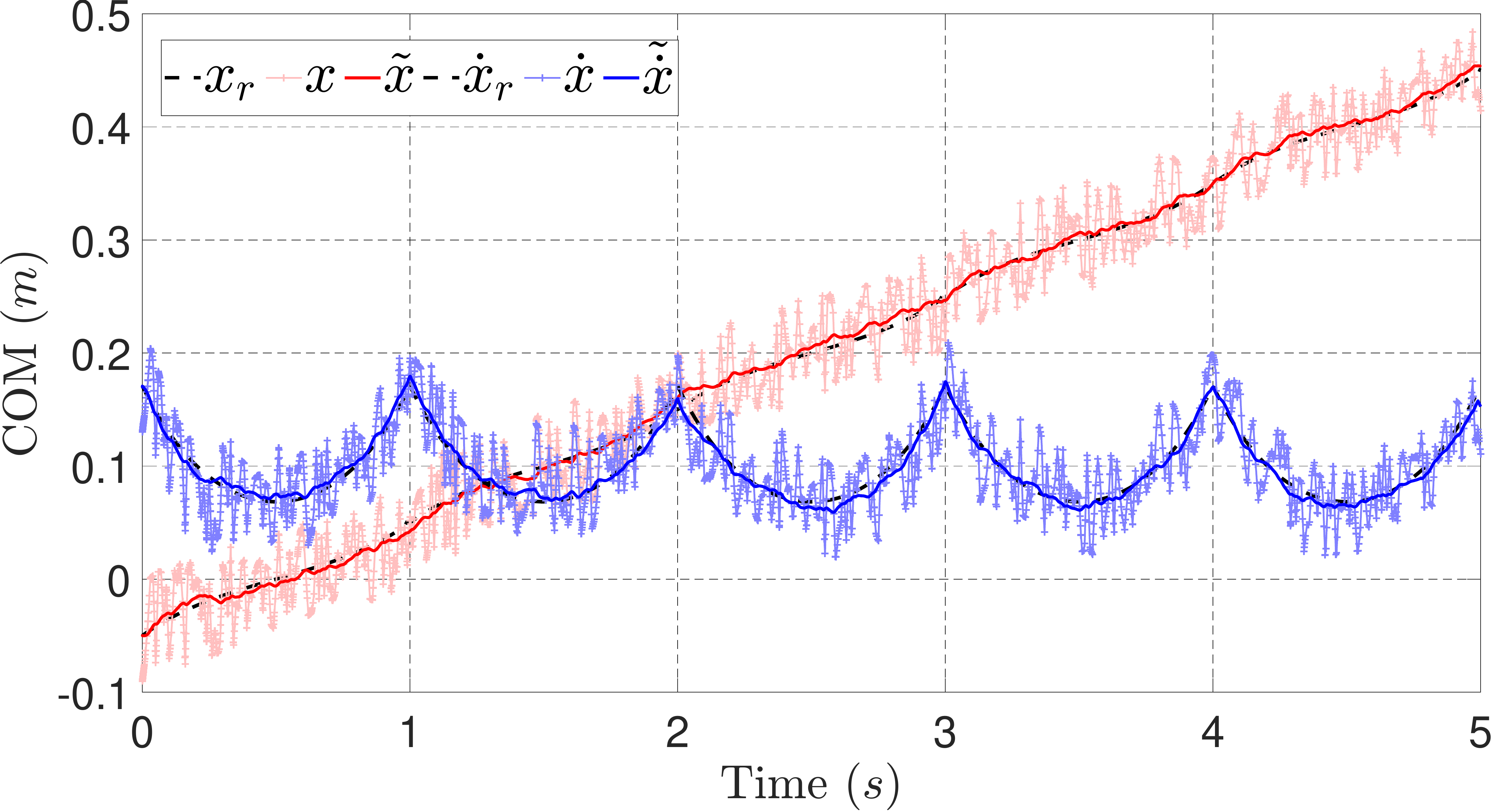} &
		\includegraphics[width=0.47\textwidth] {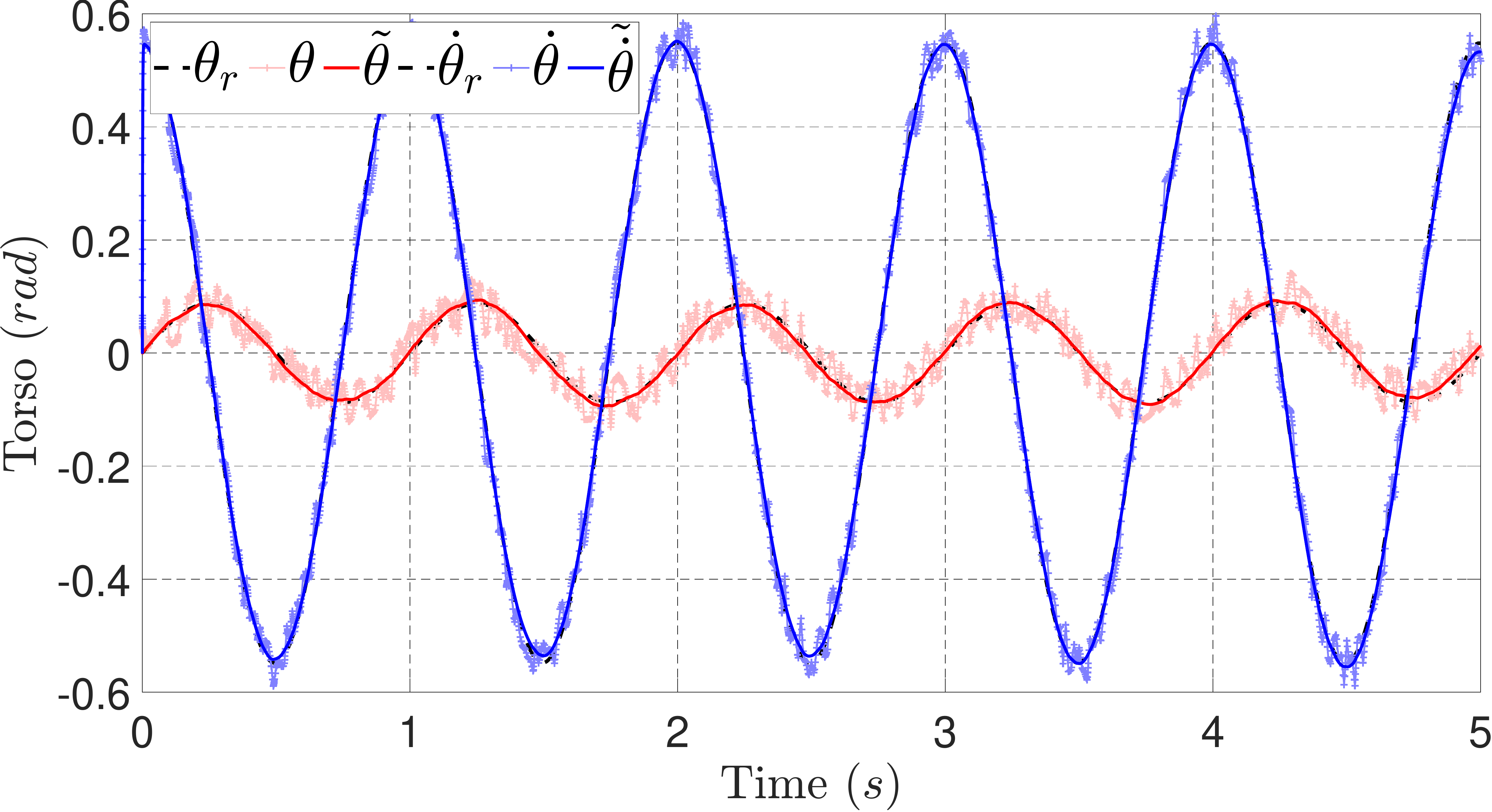}
	\end{tabular}
	\vspace{-3mm}
	\caption{Simulation results of examining the controller performance in presence of noises. In this simulation, the controller should track a references trajectories in presence of the noise$\mathcal{N}(\text{0, 6.25e-4})$.}
	\vspace{-0mm}
	\label {fig:controller_test}
\end{figure}

To check the performance of the proposed controller, a simulation has been performed. In this simulation, a set of reference trajectories has been generated and the controller should track this reference in presence of measurement noise. The simulation results are shown in Figure~\ref{fig:controller_test}. The results showed that the controller is able to track the references even when the measurements are affected by noise. In the next section, we explain how the reference trajectories are generated.

\section{Reference Trajectories Planner}
\label{sec:walking_ref}
Our walking reference trajectories planner is composed of five sub planners which are connected together hierarchically. The first level of this hierarchy is a footstep planner which generates a set of foot positions based on given step information, terrain information and some predefined constraints (e.g., maximum and minimum step length, step width, distance between feet, etc.). To do so, we consider a state variable to represent the current state of the robot’s feet:
\begin{equation}
 s = (x_l, y_l, \theta_l,\phi_l, x_r, y_r, \theta_r, \phi_r)
\end{equation}
\noindent
where $x_l, y_l, \theta_l, x_r, y_r, \theta_r$ are the position and orientation of the left and right foots, respectively. $\phi_l, \phi_r$ represent the current state of feet which is $1$ if the foot is the swing foot and $-1$ otherwise. Walking is a period motion which is generated by moving the right and the left legs alternating. Therefore, we parametrize a step action by a length and an angle from the swing foot position at the beginning of steps $a = (R, \sigma)$. According to the input parameters and the current state of the feet, an action should be taken and the state transits to a new state, $s' = t(s , a)$. Afterwards, the current footstep will be saved ($f_i\quad i \in \mathbb{N} $) and  $\phi_l$ and $\phi_r$ will be toggled. 

The second planner is the ZMP planner that uses the planned footstep information to generate ZMP reference trajectories. In our target framework, the ZMP reference planner is formulated as follows:
\begin{equation}
r_{zmp}= 
\begin{cases}
\begin{cases}
f_{i,x} \\
f_{i,y} \qquad\qquad\qquad\qquad\qquad\qquad 0 \leq t < T_{ss} \\
\end{cases} \\
\begin{cases}
f_{i,x}+ \frac{L_{sx} \times (t-T_{ss})}{T_{ds}}   \\
f_{i,y}+\frac{L_{sy}\times (t-T_{ss})}{T_{ds}} \qquad\qquad\qquad T_{ss} \leq t < T_{ss}+T_{ds} \\
\end{cases} 
\end{cases} ,
\label{eq:zmpEquation}
\end{equation}
\noindent
where $f_i = [f_{i,x} \quad f_{i,y}]$ are the planned footsteps on a 2D surface ($i \in \mathbb{N}$), $L_{sx}$ and $L_{sy}$ represent the step length and width, $T_{ss}$, $T_{ds}$ are the single support and double support durations, respectively, and $t$ is the time which will be reset at the end of each step ($t\geq T_{ss}+T_{ds}$). The third planner is the swing leg planner which generates the swing leg trajectory using a cubic spline function. This planner uses three control points that are the position of the swing leg at the beginning of the step, the next footstep position and a point between them with a predefined height~($Z_{swing}$). The fourth planner is the global sinusoidal planner which generates three sinusoidal trajectories for the COM height, the torso angles and the arm positions. The fifth planner is the hip planner which uses the generated ZMP and torso trajectories to generate hip trajectory. Indeed, these trajectories are used to determine the positions of the hip at the begging and the end of step. Using these positions, Equation~(\ref{eq:lipm_new}) can be solved as a boundary value problem as follows:
\begin{equation}
\label{eq:com_traj_x0xf}
x(t) = g_x + \frac{ (g_x-x_f) \sinh\bigl(\sqrt{\mu}(t - t_0)\bigl)+ (x_0 - g_x) \sinh\bigl(\sqrt{\mu}(t - t_f)\bigl)}{\sinh(\sqrt{\mu}(t_0 - t_f))},
\end{equation}
\noindent
where $g_x = r_{{zmp}_x}- \frac{\alpha l}{1+\alpha}\theta_{to}+\frac{\alpha\beta l}{\mu(1+\alpha\beta)}\ddot{\theta}_{to}$, $t_0$, $t_f$, $x_0$, $x_f$ are the times and corresponding positions of the hip at the beginning and at the end of a step, respectively. In this work, $T_{ds}$ is considered to be zero, which means ZMP transits to the next step at the end of each step instantaneously. Moreover, $x_f$ is assumed to be in the middle of current support foot and next support foot ($\frac{f_{i} + f_{i+1}}{2}$).

\section{Learning Framework} \label{sec:learning}

Our learning framework employs the Proximal Policy Optimization (PPO) algorithm, introduced by Schulman et al.~\cite{schulman2017ppo}, which was chosen due to its success in optimizing low-level skills concerning the NAO robot \cite{abreu2019learning,melo2019learning,teixeira2020humanoid,melo2020push,abreu2019skills}, and high-level skills \cite{muzio2020deep}, where it outperformed other algorithms such as TRPO or DDPG. The chosen implementation uses the clipped surrogate objective:

\begin{gather}
L(\theta)=\mathbb{\hat{E}}_t[min(r_t(\theta)\hat{A}_t,clip(r_t(\theta),1-\epsilon,1+\epsilon)\hat{A}_t],\nonumber\\
where \ \ r_t(\theta)=\frac{\pi_\theta(a_t | s_t)}{\pi_{\theta_{old}(a_t | s_t)}}, 
\end{gather}

\noindent where $\hat{A}_t$ is an estimator of the advantage function at timestep $t$. The clip function clips the probability ratio $r_t(\theta)$ in the interval given by $[1-\epsilon,1+\epsilon]$. This implementation alternates between sampling data from multiple parallel sources, and performing several epochs of stochastic gradient ascent on the sampled data, to optimize the objective function.

The clipping parameter $\epsilon$ was set to $0.2$, as suggested by Schulman et al.~\cite{schulman2017ppo}. Also, as in the implementations published by OpenAI for the 3D humanoid environment~\cite{baselines}, the entropy bonus was not used, and the number of optimization epochs and batches was set to 10 and 64, respectively. Some other hyperparameters were tuned using grid search: step size (\num{2.5e-4}); batch size ($4096$); and the Adaptive Moment Estimation (Adam)
~\cite{kingma2017adam} optimizer was set to use a constant scheduler. Finally, the Generalized Advantage Estimation (GAE)~\cite{schulman2018gae} algorithm's parameters --- gamma and lambda --- were set to 0.99 and 0.95, respectively, accordingly to the ranges established by the GAE's authors as best-performing for 3D biped locomotion. 

The policy is represented by a multilayer perceptron with two hidden layers of 64 neurons. The number of inputs, outputs, and the maximum number of time steps for the optimization are dependent on the scenario and will be described in section~\ref{sec:upperbody}. The training session was parallelized to improve the optimization duration.

\section{Overall Structure}
\label{sec:overall}
\begin{figure}[!t]
	\centering
    \includegraphics[width=0.99\linewidth,trim= 0cm 0cm 0cm 0cm,clip] {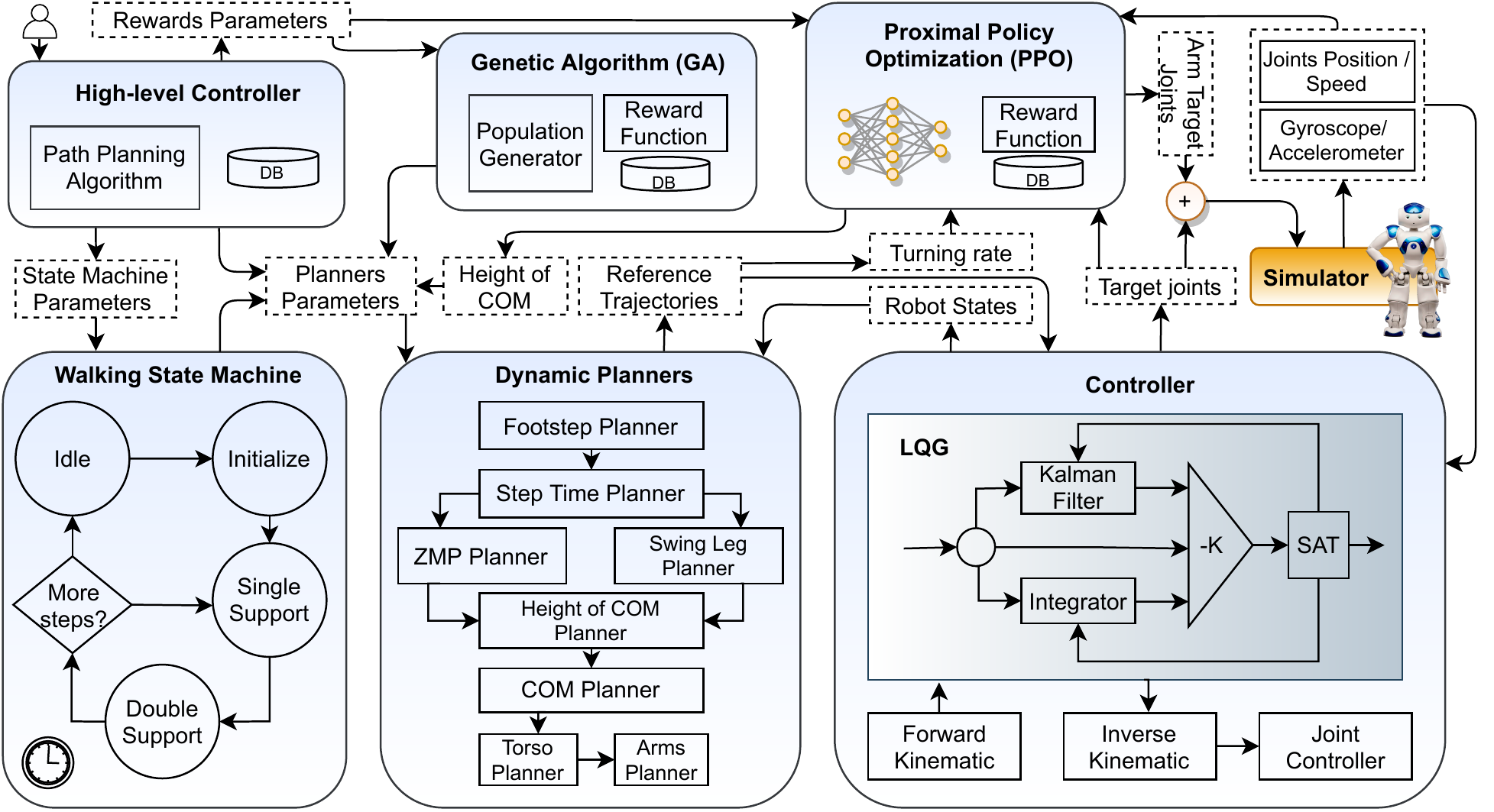}
	\caption{Overall architecture of the proposed framework. The highlighted boxes represent the main modules and the exchanged information among them is represented by the white boxes.}
	\label{fig:overall}
\end{figure}

In this section, the previously introduced planner and controller will be coupled together to generate stable locomotion. To do that, we designed a modular framework composed of six main modules. The overall architecture of this framework is depicted in Figure~\ref{fig:overall}. As shown in this figure, the walking process is controlled by a state machine which abstracts the process into four distinct states: \textit{Idle}, \textit{Initialize}, \textit{Single Support} and \textit{Double Support}. In this state machine, the transitions are triggered by a timer that is associated to each state. Additionally, it can be triggered by an emergency signal generated according to the controller's state in key moments, such as when a swift move is necessary to regain equilibrium after a strong external perturbation. The \textit{Idle state} is the initial state in which the robot is standing in place and waiting to receive a walking signal, which can be generated by an operator or a path planning algorithm. That signal triggers the \textit{Initialize state}, in which the walking parameters and configurations are loaded from a data base. Afterwards, the robot is ready to walk by shifting its COM towards the first support foot. The next state is triggered after a predefined time. During \textit{Single Support State} and \textit{Double Support State}, the dynamics planner generates the walking reference trajectories according to the generated walking signal and the controller tries to track these references.

At the same time, the neural network receives a set of observations, including data from inertial sensors, joints' position and speed, target joint positions generated by the LGQ controller, and the target turning rate. The network outputs residuals which are added to the target joint positions before being fed to the simulator, which runs the next simulation step. The neural network also adjusts the height of the COM, which is then used by the dynamic planners in the following iteration.

In the next section, a set of simulations will be carried out to verify the framework's performance. Moreover, we will show how the planners parameters are optimized using a genetic learning approach, and what is the impact of the policy gradient algorithm on the performance of the framework.

\section{Simulations}
\label{sec:Simulations}
In this section, we introduce a set of simulation scenarios to validate the performance of the proposed framework. The simulation scenarios have been designed using the official RoboCup 3D simulation environment which is based on SimSpark, a multi-agent simulator. This simulator relies on the Open Dynamics Engine~(ODE) to simulate rigid body dynamics. The physics engine is updated every 0.02s. The simulator can also be configured to update the physics engine just after receiving commands from all agents. This greatly improves simulation speed and provides a better environment for learning approaches.


\subsection{Omnidirectional Walk}

This scenario is designed to demonstrate the performance of the framework in providing an omnidirectional walk. The simulated robot starts from an idle state and follows a command comprising length~($X$), width~($Y$) and angle~($\alpha$) of the step, which is determined by an operator. Note that the step time is constant and set to 0.2s. To avoid discontinuity in the input command, a first-order lag filter is used, yielding a smooth transition. 

\begin{figure}[!t]
	\vspace{-0mm}
	\center
	\begin{tabular}	{c}	
		\includegraphics[width=0.75\textwidth , trim= 0cm 0cm 0cm 2cm,clip] {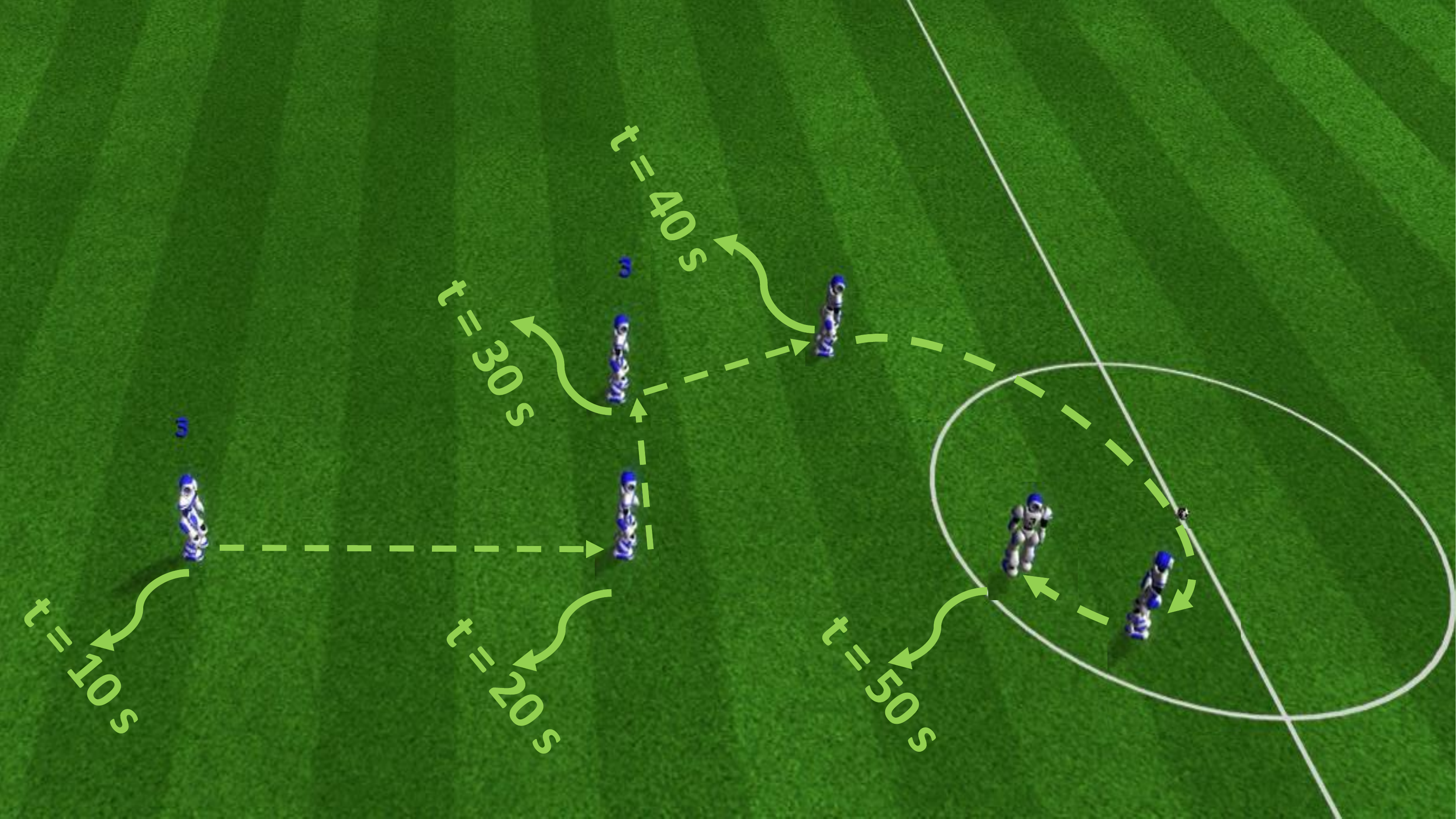}
	\end{tabular}
	\vspace{-0mm}
	\caption{ Omnidirectional walk scenario. In this scenario, the simulated robot should follow the commands that are generated by an operator: Green-dashed arrows show a set of commands that has been generated for this simulation, including forward walk, side walk, diagonal walk and turning while performing diagonal walking.}
	\vspace{-0mm}		
	\label{fig:omni_exp}	
\end{figure} 
At the beginning of this scenario, the robot is walking in place and all the setpoints are zero ($X=0.0$m, $Y=0.0$m, $\alpha=0.0\deg$/s). At $t=10$s, the operator sets the step length~($X=0.05$m) to generate forward walking;  at $t=20s$, the operator resets the step length~($X=0.0$m) and sets the step width~($Y=0.04$m) to generate side walking; at $t=30$s, while the robot is performing side walking, the operator sets the step length~($X=0.05$m) to generate diagonal walking; at $t=40$s, while the robot is performing diagonal walking, the robot is commanded to turn right simultaneously, by setting the step angle~($\alpha=10\deg$); and finally, at $t=60$s, all the set points are reset and the robot starts walking in place. A set of snapshots of this simulation is depicted in Figure~\ref{fig:omni_exp}. The simulation results showed that the framework was able to generate omnidirectional walking according to the input commands. A video of this simulation is available online at: \url{https://www.dropbox.com/s/32gml9mtumps1np/OmniWalkRC.mp4?dl=0}.

\subsection{Optimizing the Walking Parameters}
\label{sec:optimize_params}
This scenario is focused on optimizing the walking parameters to generate the fastest stable forward walk. In this scenario, the robot is placed $10$m away from the halfway line and it should walk straight forward towards that line as fast as possible. Initially, the best parameters were hand-tuned and, after several attempts, the maximum walking speed did not exceed $53$cm/s. Afterwards, based on the parametric nature of the proposed planner and controller, a GA is used to optimize the parameters to improve the walking speed. To do that, 8 parameters of the framework have been selected to be optimized. These parameters are the step length~($x$), step width~($y$), step angle~($\alpha$), height of the swing leg~($z_{sw}$), duration of a step~($T_{ss}$), torso inclination~{$TI_{to}$}, amplitude of the COM~($A_z$) and amplitude of the torso movement~($A_{to}$). In our optimization scenario, the simulated robot should walk forward for $10$ seconds and its performance will be evaluated based on the following cost function:
\begin{figure}[!t]
	\vspace{-0mm}
	\center
	\begin{tabular}	{c}	
		\includegraphics[scale=0.19, trim= 0cm 0cm 0cm 0cm,clip] {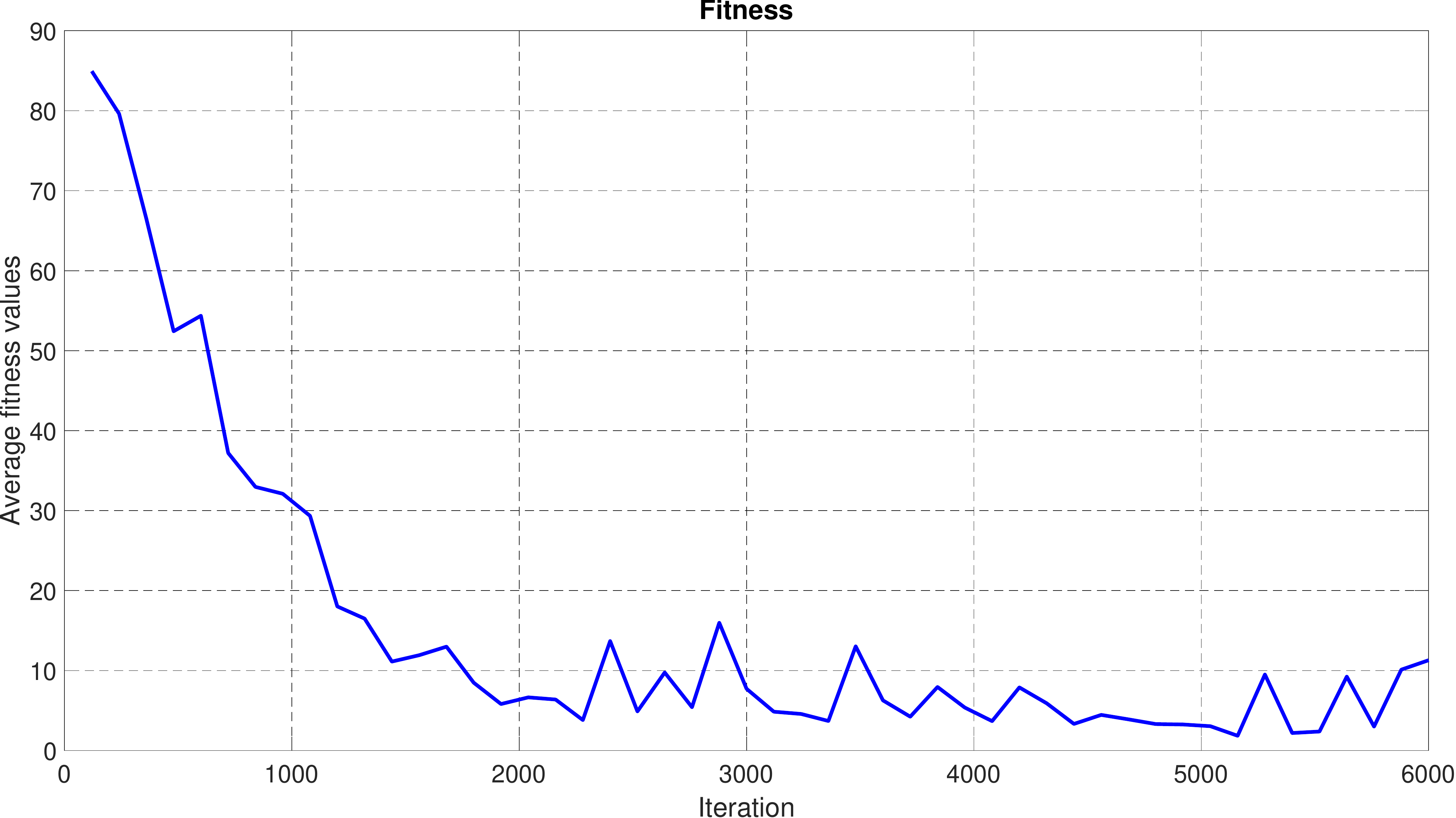}
	\end{tabular}
	\vspace{-0mm}
	\caption{ Evolution of the fitness.}
	\vspace{-0mm}		
	\label{fig:fitness}	
\end{figure} 

\begin{equation}
\label{eq:fitness}
f(\phi)= -|\delta x| + |\delta y| + \epsilon
\end{equation}
\noindent
where $\phi$ represents the selected parameters, $\delta X$, $\delta Y$ are the distance covered in~X-axis and Y-axis, respectively, $\epsilon$ is used to penalize the robot when it falls during walking~($\epsilon = 100$) otherwise it is zero. According to this cost function, the simulated robot is rewarded for straight forward walk and it is penalized for deviation and falling. A slow and stable forward walking (0.11m/s) is used as an initial solution to start the optimization process. It should be noted that, each iteration has been repeated three times and the average of the finesses was used to be sure about the walking performance. The fitness values have been recorded for each iteration and the average fitness values can be visualized in Figure~\ref{fig:fitness}. The average fitness value starts at around $85$ and after about 2000 iterations, it drops under $10$, which is much better than the first solution. The optimization has been executed for $6000$ iterations. After optimizing the parameters, the walking velocity reaches 0.866m/s, which is 61\% faster than the best hand-tuned solution.  The best parameters found by the GA are shown in Table~\ref{tb:Params}.
\begin{figure}[!t]
	\vspace{-0mm}
	\center
	\begin{tabular}	{c}	
		\includegraphics[width=0.75\textwidth , trim= 0cm 1cm 0cm 0cm,clip] {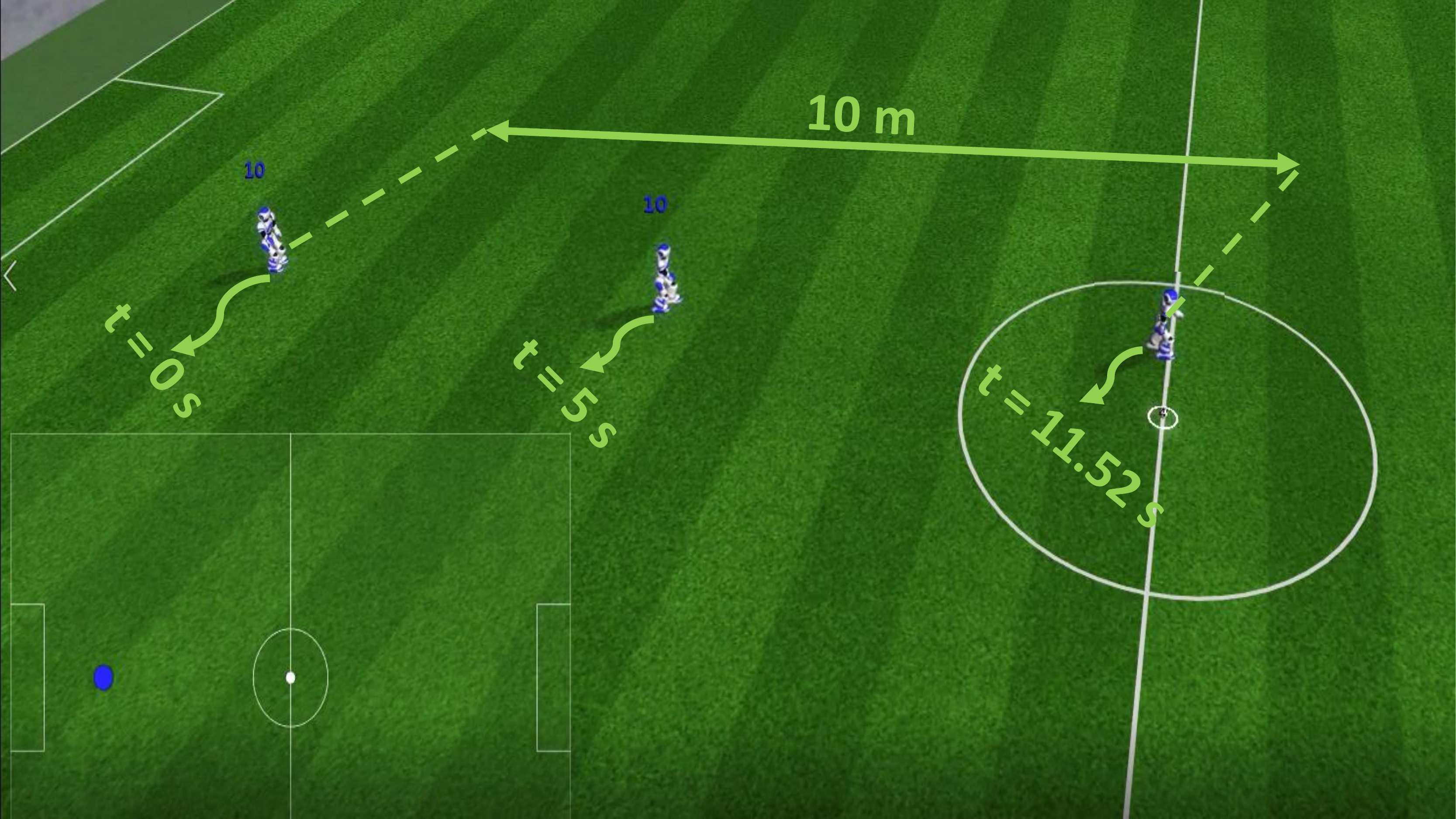}
	\end{tabular}
	\vspace{-0mm}
	\caption{ The optimization scenario and the results of an exemplary test after optimizing the parameters. In this test, the simulated robot has been placed at a specific point which is $10m$ far from the center of the field and it should walk towards the center as fast as possible. The results showed that the robot touched the midline at $t = 11.52$s.}
	\vspace{-0mm}		
	\label{fig:GA_Setup}	
\end{figure} 
\begin{table}[h]
    \small
    \centering
	\caption{The best parameters.}
	\label{tb:Params}
    \begin{tabular}{l|c|c}
        Parameter & Symbol & Value \\ \hline
        Step duration & $T_{ss}$ & 0.1274s \\ \hline
        Step length & $x$ & 0.09059m \\ \hline
        Step width & $y$ & 0.010086m   \\ \hline
        Step angle & $\alpha$ & $-0.2899\deg$ \\ \hline
        Height of swing & $z_{sw}$  & 0.038m \\ \hline
        Torso inclination &  $TI_{to}$  & $5.601\deg$ \\ \hline
        Amplitude of height of COM & $A_z$ & $-0.004$ \\ \hline
        Amplitude of torso movement & $A_{to}$ &  $-1.9195$ \\
    \end{tabular}
\end{table}
The optimization scenario and a set of snapshots of a test are shown in Figure~\ref{fig:GA_Setup}. A video of this simulated scenario is available online at: \url{https://www.dropbox.com/s/wm5y8dkekd2fnpo/OmniWalkRCOptimized.mp4?dl=0}.  

To compare the effectiveness of the dynamics model, this scenario has been repeated for the dynamics models (a) and (b) presented in Figure~\ref{fig:DynamicsModel}. To do so, the planner and the controller have been adjusted according to the dynamics models and then their parameters have been refined manually. Finally, this simulation scenario has been repeated to find the maximum forward speed of each model. The simulation results are summarized in Table \ref{tb:comparison_results_GA}. The simulation results validated that the sinusoidal motion of the height of COM improves the stability and allows the robot to move faster in comparison with fixed COM.

\begin{table}[h]
    \small
    \centering
	\caption{Summary of the results in the maximum speed scenario.}
	\label{tb:comparison_results_GA}
\begin{tabular}{c|c}
  Dynamics model                          & maximum speed \\ \hline
  LIPM                                    & 0.590 m/s     \\ \hline
  LIPM + vertical motion of COM           & 0.630 m/s     \\ \hline
  LIPM + vertical motion of COM + Torso   & 0.866 m/s     
\end{tabular}
\end{table}

    
\subsection{Learning to Improve the Upper Body Efficiency}
\label{sec:upperbody}

This scenario was designed to improve the efficiency of the walking gait in terms of speed and stability during the most common walking patterns -- forward walking and turning. To mitigate the effects of the learning process on the maneuverability and predictability of the walking trajectory, only the arms actuators and COM height were optimized.
\begin{figure}[!t]
	\vspace{-0mm}
	\center
	\begin{tabular}	{c}	
		\includegraphics[width=0.85\textwidth , trim= 0cm 0cm 0cm 0cm,clip] {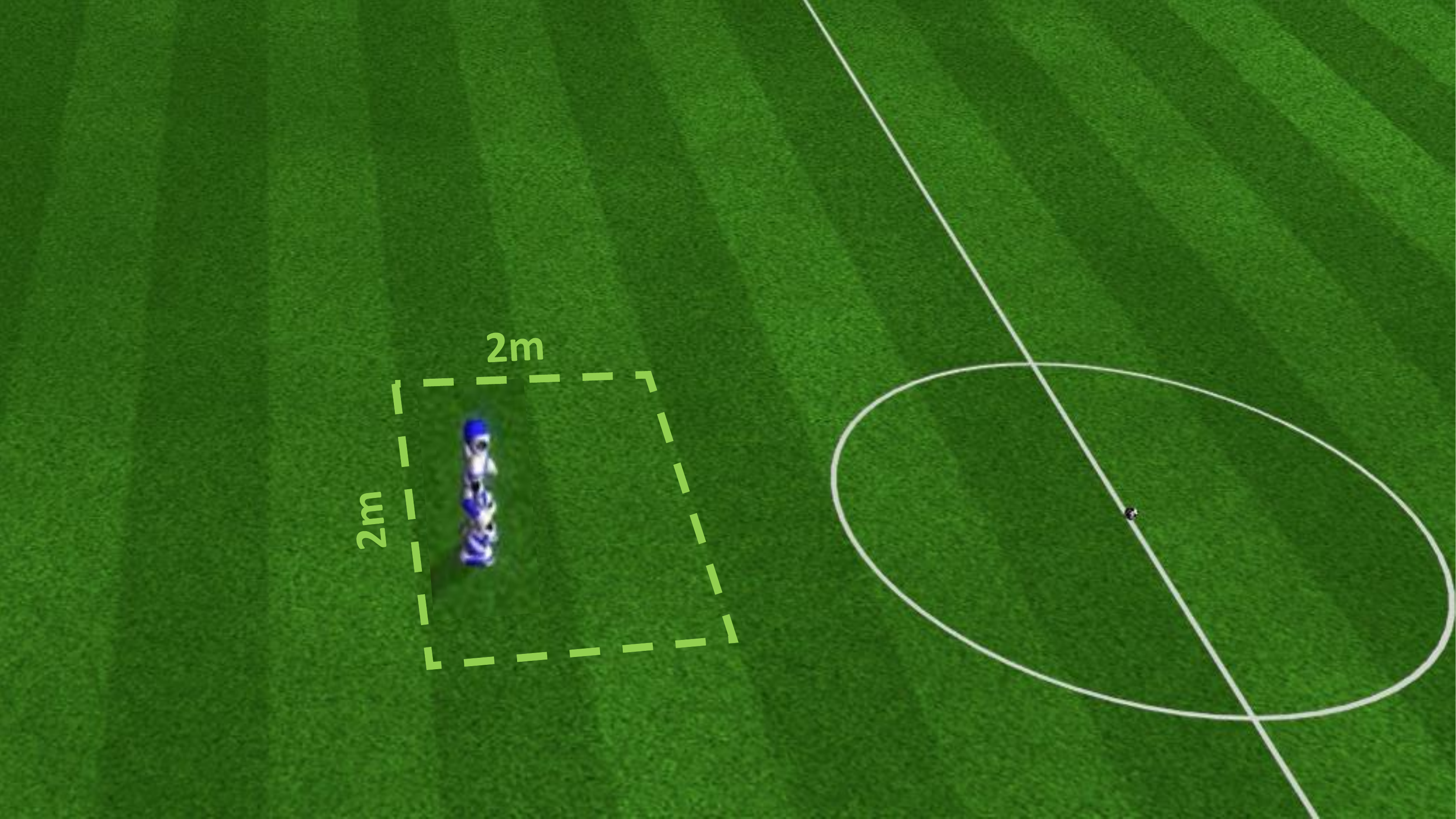}
	\end{tabular}
	\vspace{-0mm}
	\caption{ Stability optimization scenario. The robot exploits the most common walking patterns -- forward walking and turning -- to keep itself within a predefined squared area of side length 2m. When inside that area, the robot walks forward as fast as possible. Otherwise, it turns at a random rate until it is directed towards the area.}
	\vspace{-0mm}		
	\label{fig:PPO_Setup}	
\end{figure} 
The robot is initially placed in an arbitrary position within a squared area of side length 2m, as depicted in Figure~\ref{fig:PPO_Setup}. It then starts walking forward with the best parameters found in Section~\ref{sec:optimize_params}. When the robot steps out of the predefined area, it starts to turn in either direction at a random rate $|\alpha|\in[30\deg$/s$,60\deg$/s$]$, until it is facing the square again. This process is repeated continuously until the episode ends with the robot falling (detected when its $z$ coordinate drops below 0.3m). The fact that the robot runs at full speed when changing direction, and that it needs to constantly adapt to different turning rates makes this a very challenging scenario.

This optimization problem can be formalized as a Markov Decision Process (MDP) -- a 4-tuple $\left\langle S,A,p,R\right\rangle$ -- where $S$ denotes the set of states, $A$ the set of possible actions, and $R$ the set of possible numerical rewards. The dynamics of the MDP is given by the state-transition probability function $p(s'|s,a):S\times S\times A(s)\rightarrow[0,1]$ which gives the probability of ending in state $s'$ given the current state $s$ and action $a$.

The interaction between agent and environment is performed at discrete time steps ($t=0,0.02,0.04,...$). The robot's behavior was optimized by the PPO algorithm, using 67 observed variables, as listed in Table~\ref{tb:ppo_input}. The first parameter indicates the current turning rate. The inertial sensors (gyroscope and accelerometer) are both composed of 1 variable per axis in a three-dimensional space. The position and speed of all joints (excluding the head) is important to obtain a correct state representation, even though the action space only controls a limited number of these joints. Finally, the joint positions computed by the analytical controller are fed to the algorithm, and later added as residuals to the output values. These positions can be used by the network to predict the next analytical state, so that the produced residuals can be adjusted accordingly. In preliminary tests, removing this information from the state space results in a loss of performance between 5\% and 20\%, depending on possible action space combinations.

\begin{table}[!t]
    \small
    \centering
	\caption{State space for the stability optimization.}
	\label{tb:ppo_input}
    \begin{tabular}{l|c}
        Parameter & Data size ($\times$ 32b) \\ \hline
        $\alpha$ & 1 \\ \hline
        Gyroscope & 3 \\ \hline
        Accelerometer & 3 \\ \hline
        Joints Position & 20 \\ \hline
        Joints Speed & 20 \\ \hline
        Controller Actions & 20 \\ 
    \end{tabular}
\end{table}

The action space encompasses four angle variables per arm (shoulder roll, shoulder pitch, elbow yaw, elbow roll) and one variable to define the setpoint of the COM height at each step. The arm joints angles are computed by summing the analytical controller's output to the neural network's corresponding output. This forms a controller which uses the planner's arms control signals both as state data and action bias. 

The objective of this scenario is to improve forward speed and stability at all times (i.e., when moving forward or turning). The former requirement is met by rewarding the agent for stepping forward and not sideways, which can be numerically translated into the scalar projection of its velocity vector $\vec{v}$ in the direction of its orientation unit vector $\vec{o}$. Let $\vec{v}=P_t-P_{t-1}$, where $P_t$ and $P_{t-1}$ are the current and previous positions of the robot, respectively. The partial reward to motivate forward speed is then ${\max(\,\vec{v}\cdot\vec{o}\,,\,0\,)}$, where $\cdot$ denotes the dot product. The minimum reward value is limited to zero because walking backward or sideways is not worse than falling. The second requirement --- stability --- is motivated by a constant $k$, set empirically to 0.01, that rewards the agent for staying alive. More precisely, it favors stability at the cost of lowering the speed. The complete immediate reward can then be formulated as:

\begin{equation}
\label{eq:ppo_reward}
r = \max(\,\vec{v}\cdot\vec{o}\,,\,0\,)+k.
\end{equation}

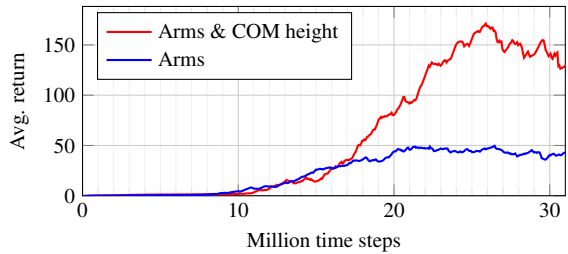
\begin{figure}[!t]
    \footnotesize
	\centering 
	\begin{tikzpicture}[]
	
	\pgfplotsset{
		height=4.1cm, width=8.0cm, compat=1.14,
	}

	\begin{axis}[
		ylabel=Avg. return,
		xlabel=Million time steps,
		legend pos=north west,
		legend cell align={left},
		ytick distance= 50,
		xtick distance= 10,
        minor x tick num=9,
        ymin=0, xmin=0, xmax=31,
        grid=both,
        minor grid style={black!5},
        major grid style={black!20},
        minor x tick style={black!10},
	]
	\addplot[red,mark=none, thick] 
	table [x=x, y=com, col sep=comma]{plots/Evo_com_nocom.csv};
	\addplot[blue,mark=none,mark options={blue, scale=0.8}, thick] 
	table [x=x, y=nocom, col sep=comma]{plots/Evo_com_nocom.csv};
	
	\legend{Arms \& COM height\\Arms\\}
	\end{axis}
	\end{tikzpicture}

	\caption{Learning curves considering only the arms (blue) and considering arms and COM height (red).}
	\label{fig:curves}
	\vspace{-0.35cm}
\end{figure}

The learning algorithm was first applied to the arms actuators and later extended to the COM height. Figure \ref{fig:curves} shows the average return evolution when learning only the arms (blue line) and after adding the COM height (red line). The former optimization plateaued at around 20M time steps, and the latter at around 26M time steps. It is important to note that the return obtained during the optimization was based on a stochastic policy whereas in the following tests, we used the corresponding deterministic policy.

The results were divided into two sections: Original scenario -- the robot is tested in the same scenario used for learning (see Figure \ref{fig:PPO_Setup}) and the analysis delves into the same metrics used to define the reward function; Straightforward path -- the robot's direction is constantly corrected to describe a linear path and the resulting behavior is analyzed in terms of efficiency.

\subsubsection{Original Scenario Results}

The robot was evaluated with regard to stability and speed in the same scenario where the learning algorithm was applied. Stability was measured by the episode length, since it terminates once the robot falls to the ground. No time limitation was imposed per episode. Speed was measured at every iteration and averaged at the end of the episode. Table \ref{tb:ppo_results} lists the average speed and duration results for 500 episodes. The first line corresponds to the walk optimized in Section \ref{sec:optimize_params}, which is used as a baseline. The robot walks on average for $5.1$ seconds with a standard deviation (SD) of 2.9s before falling, generally on the first or second sharp change of direction. The mean speed, from a stand-still position to the end of each episode, was 0.602m/s with a SD of 0.027m/s.

\begin{table}[!t]
    \small
    \centering
	\caption{Original scenario results -- average speed and duration}
	\label{tb:ppo_results}
\begin{tabular}{l|c|c}
Parameter & Ep. Length Mean $\pm$ SD (s) &  Speed Mean $\pm$ SD (m/s) \\ \hline
Baseline           & 5.1   $\pm$   2.9 & 0.602 $\pm$ 0.027 \\ \hline
Arms               & 51.6  $\pm$  31.8 & 0.710 $\pm$ 0.037 \\ \hline
Arms \& COM height & 148.2 $\pm$ 153.0 & 0.956 $\pm$ 0.060
\end{tabular}
\end{table}

After learning how to control the arms, the episode duration increased tenfold, on average, and the mean speed rose to 0.710m/s. Most falls occur at an advanced stage or during the initial sharp turns, hence the larger standard deviation. Adding the COM height to the group of controlled variables increased the episode length to almost 15 times the initial value. When compared to the version without the COM height adaptation, this metric improved approximately 3 times. The mean speed rose to 0.956m/s, a gain of almost $60\%$ in relation to the baseline. In every episode the robot eventually falls. As aforementioned, when the robot steps out of the predefined area, it starts to turn with a random turning rate between 30 and $60\deg$/s, in either direction. Most falls occur when approximating the upper limit or when the rotation direction is inverted after the robot enters and exits the predefined area in a short time (e.g. when stepping over a corner of that area). A video comparing the baseline with the Arms \& COM height optimization is available online at: \url{https://www.dropbox.com/s/d1o74tpb2u6tr9f/OmniWalkLearningComparison_subtitle.mp4?dl=0}.

\subsubsection{Straightforward Path Results}

To evaluate the gait efficiency without taking stability into account, the displacement of certain joints as well as the average speed were analyzed, as discussed later in this section. Due to the challenging nature of the learning scenario, and to provide a fair comparison with the baseline algorithm, a simplified setup was developed. The objective is to compare the baseline and best optimization algorithms while the robot tries to describe a straight path of 12m length. The turning parameter is computed at every iteration to maximize the path's linearity using a reactive proportional controller. After 12m, if the robot has not fallen, the episode is considered successful. 

\begin{figure}[!t]
    \footnotesize
	\centering 
	\begin{tikzpicture}[]
	
	\pgfplotsset{
		height=5cm, width=12.0cm, compat=1.14,
	}

	\begin{axis}[
		ylabel=Frequency,
		xlabel=Avg. speed (m/s),
		legend style={at={(0.5,0.9)},anchor=north},
		legend cell align={left},
		yticklabels={,,},
		xtick distance= 0.05,
		ytick distance= 1000,
		legend entries={Baseline,Arms \& COM height},
		major grid style={black!14},
		minor grid style={black!9},
        minor x tick style={black!20},
        grid=both,
        minor x tick num=4,
	]
	\addplot+[ybar interval, blue,mark=none,mark options={blue, scale=0.8}] 
	table [x=v1, y=freq1, col sep=comma]{plots/ppo_histogram_fwd.csv};
	\addplot+[ybar interval, red,mark=none,mark options={blue, scale=0.8}] 
	table [x=v2, y=freq2, col sep=comma]{plots/ppo_histogram_fwd.csv};
	\end{axis}
	\end{tikzpicture}
	\hfill

	\caption{Mean speed comparison between the baseline (on the left) with the best optimization using the Arms \& COM Height controller (on the right). These values were averaged for 500 successful episodes, where the robot runs for 12m in a straight line.}
	\label{fig:ppo_result_straight_speed}
\end{figure}
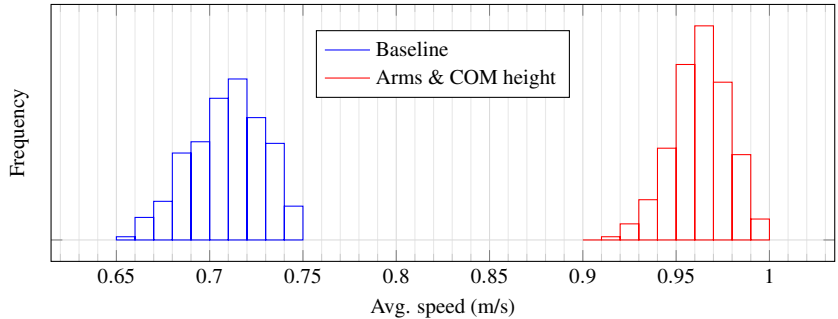

Figure~\ref{fig:ppo_result_straight_speed} compares the baseline's mean speed for the entire path with its improved version using the Arms \& COM height controller. In both cases, the speed values were averaged for 500 successful episodes. In comparison with Table \ref{tb:ppo_results}, the baseline algorithm improved its mean speed from 0.602m/s to 0.704m/s. The Arms \& COM height optimization went from 0.956m/s to 0.958m/s, indicating that the robot has a virtually constant speed, whether turning or not. The improvement of the optimized version in relation to the baseline is about $36\%$.

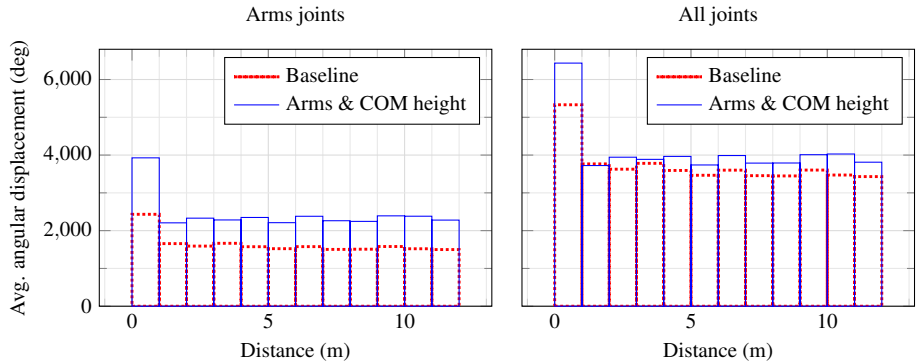
\begin{figure}[!t]
    \footnotesize
	\centering 
	\begin{tikzpicture}[]
	
	\pgfplotsset{
		height=5cm, width=6.8cm, compat=1.14
	}

	\begin{axis}[
		ylabel=Avg. angular displacement (deg),
		xlabel=Distance (m),
		legend pos=north east,
		legend cell align={left},
		ymin=0, ymax=6800,
		title=Arms joints,
		major grid style={black!14},
		minor grid style={black!9},
        minor x tick style={black!20},
        grid=both,
        minor x tick num=4,
        minor y tick num=1,
	]
	
	\addplot+[ybar interval, densely dotted, red,mark=none,line width=1.1pt] 
	table [x=d, y=e1, col sep=comma]{plots/ppo_d_en.csv};
	\addplot+[ybar interval, blue,solid,mark=none] 
	table [x=d, y=e2, col sep=comma]{plots/ppo_d_en.csv};
	\legend{Baseline\\Arms \& COM height\\}
	\end{axis}
	\end{tikzpicture}
	\hfill
	\begin{tikzpicture}[]
	
	\pgfplotsset{
		height=5cm, width=6.8cm, compat=1.14
	}
    
	\begin{axis}[
		xlabel=Distance (m),
		legend pos=north east,
		legend cell align={left},
	    ymin=0, ymax=6800,
	    yticklabels={,,},
	    title=All joints,
		major grid style={black!14},
		minor grid style={black!9},
        minor x tick style={black!20},
        grid=both,
        minor x tick num=4,
        minor y tick num=1,
	]
	
	\addplot+[ybar interval, densely dotted, red,mark=none,line width=1.1pt] 
	table [x=d, y=et1, col sep=comma]{plots/ppo_d_en.csv};
	\addplot+[ybar interval, blue,solid,mark=none] 
	table [x=d, y=et2, col sep=comma]{plots/ppo_d_en.csv};
    \legend{Baseline\\Arms \& COM height\\}
	\end{axis}
	\end{tikzpicture}

	\caption{Angular displacement performed by all arms joints (sum) during an episode, averaged for 500 successful episodes (on the left). The linear path described by the robot was divided into sections of 1m, which are represented by each bar. The baseline is represented by the dotted red bars while the optimized version is represented by the solid blue bars. The same sort of analysis for all joints is depicted on the right.}
	\label{fig:ppo_result_straight_displacement}
\end{figure}

The total angular displacement performed by certain groups of joints during a successful episode was analyzed, as this metric provides a reasonable indicator of energy consumption, considering that the actuators load is not disclosed by the server. Figure~\ref{fig:ppo_result_straight_displacement} compares the displacement sum of the arms joints with the displacement sum of all robot joints (except for the head). This analysis was performed for different stages of the linear path described by the robot. In the first meter, as expected, the robot spends more energy while gaining momentum, and then it stabilizes. Considering only the arms joints (shoulder roll, shoulder pitch, elbow yaw, elbow roll), the average angular displacement for the entire episode rose $49\%$. Despite this result, the same analysis performed for all joints yields an increment of only $10\%$. Therefore, without considering stability gains, the ratio of relative speed improvement to relative displacement increment in successful episodes is $3.6$. In essence, the robot became much more energy efficient, as a small raise in energy consumption led to a considerably faster gait.

\section{Discussion}
\label{sec:Discussion}
Simulation results showed that the framework is able to generate a fast and stable omnidirectional walk and improve its performance by learning how to control the arms and the height of the COM. Indeed, the results showed that providing a tight coupling between analytical approaches and ML improves the performance considerably. In the remaining of this section, we point out the features and limitations of the proposed framework and provide comparisons with the results of previous works.

\subsection{Features}
\begin{itemize}
\item \textbf{Architecture:} the modular architecture of the proposed framework provides some important properties such as reducing the complexity and increasing the flexibility. In comparison with approaches that are based on heuristic methods~\cite{picado2009automatic,shafii2010biped,shan2000design,lee2013generation,liu2013central,yu2013survey} or based on learning from scratch~\cite{abreu2019learning,shan2000design,endo2008learning}, our framework is able to migrate to different humanoid platforms with small changes to the control module.

\item \textbf{Computational efficiency:} unlike the approaches presented in~\cite{griffin2017walking,koryakovskiy2018model,diedam2008online,herdt2010online,griffin2016model} which are based on online optimization~(e.g., MPC), our controller was designed on top of an offline optimization algorithm. Therefore, it does not need powerful computational resources and it can be deployed on any platform easily.

\item \textbf{Considering the upper body dynamics:} most of the presented approaches in the literature used LIPM as their dynamics model, mainly due to its linearity and simplicity. Unlike LIPM-based approaches, we take into account the robot's upper body dynamics and we showed how this consideration helps to enhance the stability and speed of the robot, while improving the energy efficiency as a ratio of mean speed to total angular displacement.

\item \textbf{Release the height of COM constraint:} LIPM-based approaches assume a fixed vertical position for the COM. According to this assumption, the knee joints have to be bent while the robot is walking, which is harmful for the knee joints and causes additional energy consumption. Additionally, walking with bent knees is not very human-like. We released this constraint by assuming a sinusoidal movement for the vertical position of the COM. We showed that this assumption not only cancels the explained limitations but it also improves the stability. 

\item \textbf{Performance:} to have an entirely fair comparison, the performance of our framework should be compared with other frameworks in the same scenario and simulator. To do so, we took into consideration the maximum forward speed, and our proposed framework provides a faster walk than the agents in~\cite{picado2009automatic,asta2011nature,shafii2010biped,kasaei2017hybrid,kasaei2019fast} and slower than~\cite{macalpine2017ut,abreu2019learning,melo2019learning,abreu2019skills}. However, some of the faster examples are solely focused on sprinting forward, without the basic ability of changing direction~\cite{abreu2019learning,melo2019learning,abreu2019skills}. The comparison results are summarized in Table~\ref{tb:comparison_results}.

\begin{table}[!t]
    \small
    \centering
	\caption{Comparison Results}
	\label{tb:comparison_results}
\begin{tabular}{c|c|c}
 & Maximum speed & Ability of changing direction \\ \hline
 \cite{macalpine2017ut}     &  1.180 m/s &   YES \\ \hline
 Proposed framework         &  0.956 m/s  &  YES \\ \hline 
 \cite{kasaei2019fast}      &  0.805 m/s  &  YES \\ \hline
 \cite{shafii2010biped}     &  0.770 m/s  &  YES \\ \hline
 \cite{kasaei2017hybrid}    &  0.590 m/s  &  YES \\ \hline
 \cite{melo2019learning}    &  3.910 m/s  &  NO  \\ \hline
 \cite{abreu2019learning}   &  2.500 m/s  &  NO  \\ \hline
 \cite{abreu2019skills}     &  1.340 m/s  &  NO  \\ \hline
 \cite{asta2011nature}      &  0.550 m/s  &  NO  \\ \hline
 \cite{picado2009automatic} &  0.510 m/s  &  NO  
\end{tabular}
\end{table}

\item \textbf{Learning flexibility:} we believe that a humanoid robot should be able to learn from experience, not only to create a new behavior but also to improve its skills. Additionally, it should be able to reuse its knowledge in different scenarios. Learning how to control the arms and the COM height had a positive effect under different conditions in which the robot was not explicitly trained. The robot preserved its stability and speed when subjected to constant orientation adjustments to move in a straight line. Furthermore, we kept the learning module on top of the others to allow situations where generalization is not a conceivable solution. This is an improvement over learning from scratch approaches, as it builds upon a logical and reliable initial solution. This analytical layer is less prone to modeling errors than the learning layer, which is critical when transferring the knowledge to a real robot. After tuning the control module to new conditions, the neural network can be partially retrained by leveraging existing knowledge of similar tasks. This architecture allows for a plethora of modular optimizations aimed at stability, speed, energy efficiency, path optimization, context awareness problems (including prevention and recovery), etc. 

\item \textbf{Controller and Robustness:} some approaches~\cite{macalpine2012design,kasaei2017hybrid} used a dynamics model just to generate a feed-forward walk and did not consider any controller to track the references. Other approaches that are based on learning from scratch \cite{shan2000design,endo2008learning,abreu2019learning} do not take into account any controller explicitly. Instead, they use a learning algorithm to develop a controller implicitly. Unlike these approaches, we believe that a robust controller is an essential module of a walking framework due to the unstable nature of a humanoid robot. More specifically, when deploying the framework on a real robot, using a closed-loop walking is the best approach because it provides a better stability guarantee. Moreover, as we showed, the ML algorithms can be used on top of this controller to improve its performance. The summary of the results in the maximum speed scenario are presented in the Table~\ref{tb:comparison_results_all}.

\begin{table}[!t]
    \small
    \centering
	\caption{Summary of the results in the maximum speed scenario.}
	\label{tb:comparison_results_all}
\begin{tabular}{c|c|c}
  Dynamics model                           & ML algorithm  & maximum speed \\ \hline
  LIPM                                     & GA            & 0.590 m/s     \\ \hline
  LIPM + vertical motion of COM            & GA            & 0.630 m/s     \\ \hline
  LIPM + vertical motion of COM+ Torso     & GA            & 0.866 m/s     \\ \hline
  LIPM                                     & GA+PPO        & 0.710 m/s     \\ \hline
  LIPM + vertical motion of COM            & GA+PPO        & 0.741 m/s     \\ \hline
  LIPM + vertical motion of COM+ Torso     & GA+PPO        & 0.956 m/s     \\ \hline
  
\end{tabular}
\end{table}

    
\end{itemize}

\subsection{Limitations}
\begin{itemize}
    \item \textbf{Swing leg dynamics:} the legs of a humanoid robot are generally composed of six joints and have non-negligible masses. In our dynamics models, the swing leg is considered to be massless, which affects the controller performance. Taking into account the inertia and mass of the swing leg can minimize tracking errors and improve the controller's performance~\cite{faraji20173lp,kasaei2020robust}.
    \item \textbf{Reality Gap:} the disparity between reality and simulation is a matter of concern when employing offline ML techniques. Learning to improve the upper body efficiency took between 20 and 26 million time steps. Other works have shown the optimization of robotic tasks, such as squatting \cite{koryakovskiy2018model}, using RL combined with an analytical controller, in under 10M time steps. Haarnoja et al. also demonstrated that learning humanoid tasks from scratch can also be performed in about the same period of time \cite{haarnoja2018soft}. However, it must be noted that these approaches employed distinct environments with different robots, directly influencing the complexity of the task. Learning to run using the NAO robot in SimSpark can take close to 200M time steps \cite{melo2019learning,abreu2019learning}. Nevertheless, 20-26 million time steps can still be characterized as poor sample efficiency, as it takes a considerable amount of time and must be performed in a simulated environment. The gap between both worlds largely affects the transferability of knowledge to the real robot. Despite the scientific community's considerable effort to reduce this gap, it remains an issue when dealing with intricate robot models. Additionally, it is not possible to learn directly on the real robot due to the high potential of mechanical damage.   
\end{itemize}

\section{Conclusion}
\label{sec:Conclusion}
In this paper, we have tackled the problem of developing a robust biped locomotion framework by proposing a tight coupling between an analytical control approach and a reinforcement learning approach. The overall architecture of the framework was composed of six distinct modules which were hierarchically structured. We abstracted the overall dynamics of a humanoid robot into two masses. Then, we used the ZMP concept and some assumptions to represent this dynamics model as a linear state space system. The system was composed of four states and we explained how it can be used to plan and control the walking reference trajectories. Particularly, the planner was composed of five sub planners and the controller was formulated as an LQG controller, which is not only robust against uncertainties but also provides a promising solution using an offline optimization. We analyzed the performance of the controller in the presence of uncertainties using simulations and the results validated its performance. Moreover, we illustrated how the parametric nature of the framework allows us to use the PPO algorithm on top of an analytical control approach to improve the performance of the framework. Finally, the performance of the proposed framework was validated in several simulated scenarios. The first two scenarios were focused on examining the ability of the framework in generating an omnidirectional walk and finding the maximum velocity of the forward walk. The third scenario was designed to assess the capability of the learning module in improving the framework's performance. The robot learned how to move its arms and COM height in order to improve the stability, speed and energy efficiency. This limited action space enabled the robot to learn how to walk without falling for much longer periods (almost 15 times longer), while also improving the speed by $60\%$ when walking forward or turning. 

As future work, we would like to design a more accurate dynamics model by considering the mass of the swing leg to improve the framework's performance. Additionally, we would like to extend our framework by adding another module to handle the emergency conditions based on learning a set of specific actions. 

\section*{Acknowledgement}
This research is supported by Portuguese National Funds through Foundation for Science and Technology (FCT) through FCT scholarship SFRH/BD/118438/2016. The second author is supported by FCT under grant SFRH/BD/139926/2018.

\bibliographystyle{elsarticle-num}
\bibliography{RAS_PPO}

\end{document}